\documentclass{article}

\usepackage{amsmath,amsfonts,bm}

\def\eqref#1{equation~\ref{#1}}
\def\1{\bm{1}}

\def\mW{{\mathbf{W}}}

\def\mmu{{\bm{\mu}}}
\def\mSigma{{\bm{\Sigma}}}

\def\mtheta{{\bm{\theta}}}

\def\vw{{\bm{w}}}
\def\vx{{\bm{x}}}
\def\vy{{\bm{y}}}
\def\vz{{\bm{z}}}

\def\mW{{\bm{W}}}

\def\mSigma{{\bm{\Sigma}}}

\DeclareMathAlphabet{\mathsfit}{\encodingdefault}{\sfdefault}{m}{sl}
\SetMathAlphabet{\mathsfit}{bold}{\encodingdefault}{\sfdefault}{bx}{n}

\def\gB{{\mathcal{B}}}

\def\gD{{\mathcal{D}}}

\def\gH{{\mathcal{H}}}

\def\gL{{\mathcal{L}}}

\def\gN{{\mathcal{N}}}

\newcommand{\E}{\mathbb{E}}

\newcommand{\KL}{D_{\mathrm{KL}}}

\DeclareMathOperator*{\argmax}{arg\,max}
\DeclareMathOperator*{\argmin}{arg\,min}

\usepackage{microtype}
\usepackage{graphicx}
\usepackage{subcaption}
\usepackage{booktabs} %
\usepackage{multirow,multicol}
\usepackage[colorinlistoftodos,prependcaption,textsize=tiny]{todonotes}

\renewcommand{\eqref}[1]{(\ref{#1})}
\usepackage{hyperref}

\usepackage{standalone}
\usepackage{tikz}
\usetikzlibrary{positioning}
\usetikzlibrary{arrows}
\usetikzlibrary{calc,fit}
\usetikzlibrary{shapes.geometric}
\usetikzlibrary{shapes.misc}
\usetikzlibrary{decorations.pathmorphing}
\usetikzlibrary{decorations.pathreplacing}
\usepackage{pgfplots}
\pgfplotsset{compat=1.17}

\usepackage[preprint]{icml_style}

\usepackage{amsmath}
\usepackage{amssymb}
\usepackage{mathtools}
\usepackage{amsthm}
\usepackage{bbm}

\usepackage{xspace}

\usepackage{newtxmath}

\usepackage[nameinlink,capitalize,noabbrev]{cleveref}

\theoremstyle{plain}

\theoremstyle{definition}

\theoremstyle{remark}

\usepackage[textsize=tiny]{todonotes}

\newcommand{\sstd}[1]{\textcolor{black}{\tiny{$\pm #1$}}}
\newcommand{\highlight}[1]{\colorbox{blue!10}{#1}}
\newcommand{\norm}[1]{\left\lVert#1\right\rVert}

\newcommand{\iid}{\textit{i.i.d.}\xspace}
\newcommand{\ie}{\textit{i.e.}}
\newcommand{\eg}{\textit{e.g.}}

\newcounter{daggerfootnote}
\newcommand*{\daggerfootnote}[1]{%
    \setcounter{daggerfootnote}{\value{footnote}}%
    \renewcommand*{\thefootnote}{\fnsymbol{footnote}}%
    \footnote[2]{#1}%
    \setcounter{footnote}{\value{daggerfootnote}}%
    \renewcommand*{\thefootnote}{\arabic{footnote}}%
    }

\makeatletter
\renewcommand{\paragraph}[1]{\textbf{#1}}
\makeatother

\icmltitlerunning{In-Context Parametric Inference: Point or Distribution Estimators?}

\begin{document}

\twocolumn[
\icmltitle{In-Context Parametric Inference: Point or Distribution Estimators?}

\icmlsetsymbol{equal}{*}

\begin{icmlauthorlist}
\icmlauthor{Sarthak Mittal}{udem,mila}
\icmlauthor{Yoshua Bengio}{udem,mila}
\icmlauthor{Nikolay Malkin}{edinburgh}
\icmlauthor{Guillaume Lajoie}{udem,mila}
\end{icmlauthorlist}

\icmlaffiliation{udem}{Universit\'e de Montreal}
\icmlaffiliation{mila}{Mila}
\icmlaffiliation{edinburgh}{University of Edinburgh}

\icmlcorrespondingauthor{Sarthak Mittal}{sarthmit@gmail.com}

\icmlkeywords{Machine Learning, ICML}

\vskip 0.3in
]

\printAffiliationsAndNotice{} %

\begin{abstract}
Bayesian and frequentist inference are two fundamental paradigms in statistical estimation. Bayesian methods treat hypotheses as random variables, incorporating priors and updating beliefs via Bayes' theorem, whereas frequentist methods assume fixed but unknown hypotheses, relying on estimators like maximum likelihood. While extensive research has compared these approaches, the frequentist paradigm of obtaining point estimates has become predominant in deep learning, as Bayesian inference is challenging due to the computational complexity and the approximation gap of posterior estimation methods. However, a good understanding of trade-offs between the two approaches is lacking in the regime of amortized estimators, where in-context learners are trained to estimate either point values via maximum likelihood or maximum a posteriori estimation, or full posteriors using normalizing flows, score-based diffusion samplers, or diagonal Gaussian approximations, \textit{conditioned} on observations. To help resolve this, we conduct a rigorous comparative analysis spanning diverse problem settings, from linear models to shallow neural networks, with a robust evaluation framework assessing both in-distribution and out-of-distribution generalization on tractable tasks. Our experiments indicate that amortized point estimators generally outperform posterior inference, though the latter remain competitive in some low-dimensional problems, and we further discuss why this might be the case.\daggerfootnote{Official code for the work can be found \href{https://github.com/sarthmit/parametric_inference}{here}.}
\end{abstract}

\section{Introduction}

Bayesian and requentist inference represent two core principles to statistical estimation
and machine learning that provide complementary approaches to model training and evaluation. Bayesian methods treat hypotheses $\theta$, such as model parameters, as random variables and use data $\gD$ as evidence to update posterior beliefs $p(\theta\mid\gD)$, whereas frequentist methods, such as maximum likelihood and moment methods, assume fixed but unknown hypotheses $\theta^*$ and estimate them through optimization. Despite the dominance of the Bayesian approach in the earliest successes of generative modeling \citep[][among many others]{hinton1995wake,neal1996bayesian}, the frequentist paradigm of obtaining point estimates has become predominant in deep learning, as Bayesian inference is challenging due to the complexity of estimating the posterior distribution \citep{blei2017variational}.

An understanding of trade-offs between the two approaches, and between different methods for posterior approximation, is lacking in the regime of amortized estimators \citep{kingma2013auto,rezende2014stochastic,garnelo2018neural}, where models $q_\phi(\theta\mid\gD)$ that take the dataset $\gD$ explicitly as input are trained to estimate either point values or parametric distributions over $\theta$. The Bayesian posterior predictive minimizes empirical risk, and it should be optimal to use it in prediction problems compared to a point estimate \citep{devroye1996probabilistic}. However, this optimality may not hold when approximate families that cannot express the full posterior are used, or when an amortization gap is introduced by a model that takes data as input and must generalize to new observations \citep{cremer2018inference} in-context.

The consequences of such limitations of amortized inference have been noted in a sequence of works in diverse areas of deep learning. For instance, higher variational bounds on likelihood do not necessarily lead to better models in VAEs \citep{rainforth2018tighter} or to more effective approximations to the target distribution in variational methods \cite{blessing2024beyond}. Similarly, for Bayesian neural networks (BNNs), the effectiveness of simple approximating families for the posterior compared to methods like MCMC has been debated \citep{ritter2018scalable}; indeed, posteriors need to be integrated at lower temperatures to useful approximate posterior predictive distributions \citep{wenzel2020good,adlam2020cold}. These findings are also relevant to recent work on in-context learning in large language models, which approximates Bayesian inference at optimality \citep{xie2021explanation,akyurek2022learning} but falls short in practice \citep{garg2022can,falck2024incontext}.

In this paper, we conduct a comparative analysis on several problem settings, from linear models to shallow neural networks, to assess the performance of different inference methods in both in-distribution (ID) and out-of-distribution (OoD; misspecified) settings. We compare various generative modeling, variational and point estimation algorithms for inferring underlying parameters (\cref{fig:figure1}). Our experiments indicate that amortized point estimators generally outperform Bayesian methods, especially on high-dimensional tasks. Our results contribute evidence from simple, well-understood models and inference procedures to the debate on the utility of approximate Bayesian inference in deep learning, especially in an \textit{in-context} setting.

\section{Problem Setup}

We consider a generative model of sets of \iid observations $\gD = \{(\vx_i, \vy_i)\}_{i=1}^k$, assumed to be sampled from a ground-truth underlying distribution: $(\vx_i,\vy_i)\sim\chi$. Given a parametric family of conditional models $p(\vy \mid \vx, \theta)$, we would like to infer the parameter $\theta$ that best explains the data $\gD$. Inferring $\theta$ allows us to make predictions of $\vy$ on new data points $\vx$, by computing $p(\vy\mid\vx,\theta)$. 

If the true model $p(\vy\mid\vx)$ -- the conditioning of $\chi$ on $\vx$\footnote{We use distributions and probability or mass functions interchangeably and assume that disintegration of the joint distribution is possible, as we consider only variables valued in $\mathbb{R}^n$ with absolutely continuous densities and in discrete spaces.} -- lies in the model class considered (that is, it is equal to $p(\vy\mid\vx,\theta^*)$ for some $\theta^*$), then we hope to recover the parameter $\theta^*$, or a model equivalent to it. If the true model does not lie in the model class considered, the inference problem is said to be misspecified.

There are two main paradigms for estimating $\theta$ from $\gD$: frequentist and Bayesian. Frequentist methods treat $\theta$ as fixed but unknown and estimate it by optimizing a functional that is maximized when $\theta=\theta^*$. Bayesian methods treat $\theta$ as a random variable and approximate a distribution over it by positing a prior distribution $p(\theta)$ and matching, by various means, the posterior distribution $p(\theta\mid\gD)$. 

Both approaches primarily operate on a fixed set of observations $\gD$ and rely on iterative methods to infer $\theta$. In this work we are interested in in-context estimators\footnote{We use \emph{in-context} and \emph{amortized} estimators interchangeably.} that explicitly take $\gD$ as input and provide an estimate for $\theta$, and can generalize to new datasets zero-shot.

We briefly review the frequentist and Bayesian methods below, with a focus on amortized estimators.

\subsection{Frequentist Estimation}

The most common frequentist estimator is the maximum likelihood estimator (MLE), which estimates $\theta$ by maximizing the likelihood of the data $\gD$ under the model $p(\vy\mid\vx,\theta)$. The MLE is defined as
\begin{equation}\label{eq:mle}
    \theta_{\rm MLE} 
    = \argmax_\theta p(\gD\mid\theta)
    = \argmax_\theta \sum_{i=1}^k \log p(\vy_i\mid\vx_i,\theta),
\end{equation}
supposing this optimum exists. Other frequentist estimators include moment methods \citep{pearson1936method}. An in-context version of frequentist estimators can be seen as
\begin{align}
f(\gD; \phi^*) \approx \theta_{MLE}(\gD) \qquad\text{where}\quad \gD\sim\chi
\end{align}
where the parameters $\phi^*$ can analogously be estimated as
\begin{align}
    \phi^* = \arg\max_\phi \mathbb{E}_{\gD \sim \chi}  \log p(\gD | \theta = f(\gD; \phi))
\end{align}

\subsection{Bayesian Estimation}

Given a prior distribution $p(\theta)$, we have the posterior distribution
\begin{equation}\label{eq:bayes}
    p(\theta\mid\gD) 
    \propto p(\gD\mid\theta)p(\theta)
    = p(\theta)\prod_{i=1}^k p(\vy_i\mid\vx_i,\theta).
\end{equation}
The simplest Bayesian estimator is the a point estimate of the mode of $p(\theta\mid\gD)$, called the maximum a posteriori (MAP) estimator:
\begin{align}\nonumber
    \theta_{\rm MAP} 
    &= \argmax_\theta p(\theta\mid\gD)\\\label{eq:map}
    &= \argmax_\theta \log p(\theta) + \sum_{i=1}^k \log p(\vy_i\mid\vx_i,\theta)
\end{align}
(note the similarity with \eqref{eq:mle}). \emph{Posterior concentration} results (\citet{doob1949}; see also \citet{miller2018detailed,miller2021asymptotic}) show that, under some conditions, the posterior distribution concentrates around the true parameter value $\theta^*$ as $k\to\infty$, meaning that prior term in \eqref{eq:map} becomes irrelevant and the MAP and MLE converge to the same value. Such results hold almost surely with respect to the \iid sampling of data from the true distribution $\chi$ and assume the model class $p(\vy\mid\vx,\theta)$ contains the true model.

While the MAP estimator approximates the posterior $p(\theta\mid\gD)$ in \eqref{eq:bayes} by a point mass at $\theta_{\rm MAP}$, other Bayesian methods may approximate it by more complex distributions, such as parametric models $q_\phi(\theta)$. The goal is to infer parameters $\phi^*$ that bring $q_\phi$ close to the true posterior in some measure of divergence $\mathbb{D}$,
\begin{align}
    \phi^* = \arg\min \mathbb{D}\left(p(\cdot | \gD), q_\phi(\cdot)\right)
\end{align}
When $q_\phi$ is a model taking $\gD$ explicitly as input, it is called an \emph{amortized} estimator. The goal of amortized inference is to learn a model $q$ that can approximate the true posterior in a fast and scalable manner. 

Amortized estimators are the focus of this work. The parametrization of amortized inference models will be discussed in \cref{sec:amortization_icl} and training objectives in \cref{sec:training_objectives}.

\begin{figure}
    \centering
    \begin{tikzpicture}[
    every node/.style={inner sep=0pt},
    algo/.style={},
    group/.style={draw, rounded corners=6pt, thick, inner sep=6pt, minimum width=130pt},
    crossgroup/.style={draw, rounded corners=6pt,inner sep=8pt, very thick, dashed},
    biggroup/.style={draw, rounded corners=6pt, very thick, inner sep=6pt, minimum width=140pt},
    y=18pt, 
    x=150pt]

\definecolor{bayescolor}{HTML}{9b59d0}
\definecolor{freqcolor}{HTML}{ffe433}

\pgfdeclarelayer{background}
\pgfsetlayers{background,main}

\node[algo] (algo_gaussian_forward) at (0,0) {Gaussian (Forward KL)};
\node[algo] (algo_nf_forward) at (0,-1) {Norm.\ Flow (Forward KL)};
\node[algo] (algo_dsm) at (0,-2) {Diffusion (Score Matching)};
\node[algo] (algo_fm) at (0,-3) {CNF (Flow Matching)};
\node[group, fit=(algo_gaussian_forward) (algo_nf_forward) (algo_dsm) (algo_fm)] (group_sample) {};

\node[algo] (algo_gaussian_sym) at (0,-5.5) {Gaussian (Symmetric KL)};
\node[algo] (algo_nf_sym) at (0,-6.5) {Norm.\ Flow (Symmetric KL)};
\node[group, fit=(algo_gaussian_sym) (algo_nf_sym)] (group_sym) {};

\node[algo] (algo_gaussian_reverse) at (0,-9) {Gaussian (Reverse KL)};
\node[algo] (algo_nf_reverse) at (0,-10) {Norm.\ Flow (Reverse KL)};
\node[algo] (algo_dem) at (0,-11) {Diffusion (DEM)};
\node[algo] (algo_map) at (0,-12) {MAP};
\node[group, fit=(algo_gaussian_reverse) (algo_nf_reverse) (algo_dem) (algo_map)] (group_variational) {};

\node[algo] (algo_mle) at (-1,-12) {MLE};

\node[above=3pt of group_sample] (group_sample_title) {\it Sample-based};
\node[above=3pt of group_sym] {\it Sample-based + Variational};
\node[above=3pt of group_variational] {\it Variational\vphantom{p}};

\node[group, fit=(algo_mle), draw=none] (group_frequentist) {};

\begin{pgfonlayer}{background}
    \node[biggroup, fill=freqcolor!10, fit=(group_frequentist)] (biggroup_frequentist) {};
    \node[biggroup, fill=bayescolor!10, fit=(group_sample_title) (group_sym) (group_variational)] (biggroup_bayesian) {};
\end{pgfonlayer}

\node[below=6pt of biggroup_frequentist] {\bf\large Frequentist};
\node[below=6pt of biggroup_bayesian] {\bf\large Bayesian};

\node[crossgroup, fit=(algo_map) (algo_mle),yshift=-1pt,draw=red] (group_point) {};
\node[below=6pt of group_point] {\it\color{red} Point estimation};

\node[algo] (prior) at (-1,-0) {$p(\theta)$, $p(\mathbf{y}\mid\mathbf{x},\theta)$};
\node[algo] at (-1,1) {\bf Prior and model specification};

\node[algo] (dataset) at (-1,-2) {$\mathcal{D}=\left\{(\mathbf{x}_i,\mathbf{y}_i)\right\}$};
\node[algo] (dataset_iid) at (-1,-3) {$\theta\sim p(\theta)$, $\mathbf{y}_i\sim p(\mathbf{y}_i\mid \mathbf{x}_i,\theta)$};
\node[algo] at (-1,-1) {\bf Dataset (well-specified)};

\node[algo] (likelihood) at (-1,-5) {$p(\mathcal{D}\mid\theta)=\prod_i p(\mathbf{y}_i\mid\mathbf{x}_i,\theta)$};
\node[algo] (posterior) at (-1,-6) {$p(\theta\mid\mathcal{D})\propto p(\mathcal{D}\mid\theta)p(\theta)$};
\node[algo] at (-1,-4) {\bf Bayesian posterior};

\node[algo] (varposterior) at (-1,-8) {$q_\phi(\theta\mid\mathcal{D})$};
\node[algo] at (-1,-7) {\bf Variational posterior};

\node[algo] (postpred) at (-1,-10) {$p(\mathbf{y}\mid\mathbf{x},\mathcal{D})\approx\!\!\operatorname*{\mathbb{E}}\limits_{q_\phi(\theta\mid\mathcal{D})}[p(\mathbf{y}\mid\mathbf{x},\theta)]$};
\node[algo] at (-1,-9) {\bf Posterior predictive};

\end{tikzpicture}
    \vspace{-1mm}
    \caption{A hierarchical decomposition of the suite of in-context estimators which we compare in this work. Each estimator explicitly looks at the observations as input and optionally takes the prior into account to provide either a single point estimate, or a distribution, for the parameters. We consider three broad classes of Bayesian methods:
    Sample-based, which require access to simulated data, Variational, which require access to the joint density, and those that require both.}
    \label{fig:figure1}
    \vspace{-5mm}
\end{figure}
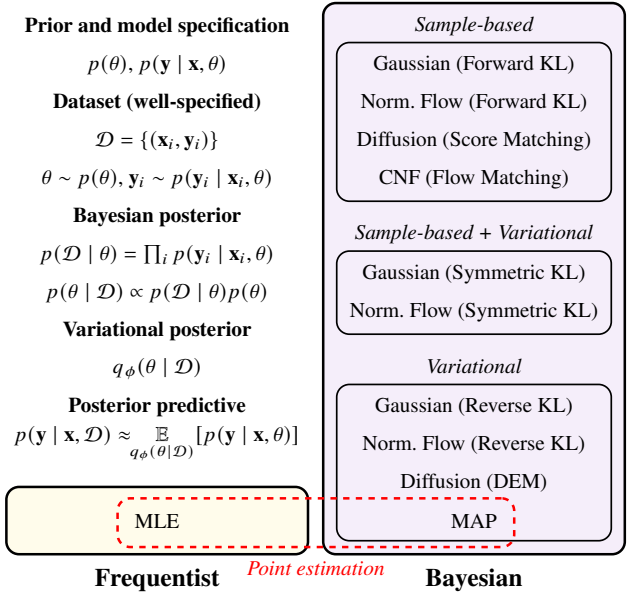
\subsection{Posterior Predictive Distributions} 

Once $\theta$ is estimated by a distribution $q_\phi(\theta)$ we can make predictions on new data points $\vx$ by computing the posterior predictive distribution
\begin{align}\nonumber
    p(\vy\mid\vx,\gD) 
    &= \int p(\vy\mid\vx,\theta)p(\theta\mid\gD)\,d\theta \\\label{eq:postpred}
    &\approx \E_{\theta\sim q_\phi(\theta)} p(\vy\mid\vx,\theta).
\end{align}
For point estimates, we can use the MAP or MLE estimate of $\theta$ in place of the expectation in \eqref{eq:postpred}. 

The main question we address is: which estimates of the model parameters $\theta$ give the best predictions on new data points $\vx$ via the posterior predictive \eqref{eq:postpred}?

\subsection{Amortization by In-Context Estimation}

\label{sec:amortization_icl}
Traditionally, in-context learning (ICL)~\citep{dong2022survey} over a training set $\gD$ refers to the ability of a pretrained sequence model (\eg, a LLM) to solve novel tasks when presented with the examples from $\gD$ in-context. A number of works~\citep{akyurek2022learning,garg2022can,xie2021explanation,von2023transformers,mittal2024does,elmoznino2024context} formalize this form of ICL from the perspective of algorithmic tasks as directly modeling the posterior predictive model $\arg\max_\phi \mathbb{E}_{\vx, \vy, \gD \sim \chi}\log p_\phi(\vy | \vx, \gD)$, where $\chi$ defines some data distribution. 

While ICL methods often model the posterior predictive, they can be adapted to perform parametric inference given a likelihood function \citep{mittal2023exploring}. Generally, parametric inference relies on approximate procedures to obtain estimates for a particular task and set of observations -- \eg~ MLE of neural network parameters relies on gradient descent or posterior samples through MCMC approaches, for a fixed set of observations $\gD$ (training set). Instead, we are interested in amortized / in-context estimators that explicitly take the observations as input and output the estimates, whether probabilistic or point. Such an in-context estimator's task is to model parameter inference, as opposed to prediction, and can provide estimates for novel tasks in zero-shot and compute-efficient manner. We rely on a transformer architecture to model conditioning on $\gD$ and omit positional embeddings to satisfy permutation invariance of the posterior given \iid samples.

\section{Amortized Inference Training Objectives}

\label{sec:training_objectives}
We formalize different training objectives and parametrizations for learning amortized estimators to approximate either point estimates or full posteriors. Within posterior estimation, we consider three classes of algorithms based on the learning signal used: sample-based, variational methods or a combination of the two. Refer to \cref{fig:figure1} for a hierarchical view over the different in-context estimators considered.

\subsection{Point Estimates}

For the point estimates $\theta_{\rm MLE}$ and $\theta_{\rm MAP}$, an amortized model $\theta=f(\gD;\phi)$ directly outputs the parameter value $\theta$ given the data $\gD$. The optimization problems in \eqref{eq:mle} and \eqref{eq:map} can be directly used as the objectives for training $\phi$; for example, for MAP:
\begin{equation}\label{eq:mle_loss}
    \gL_{\rm MAP}(\gD) = \log p(f(\gD;\phi)) + \sum_{i=1}^{|D|} \log p(\vy_i\mid\vx_i,f(\gD;\phi)).
\end{equation}
An unbiased estimator of the gradient of $\gL_{\rm MAP}(\gD)$ can be obtained by a stochastic surrogate loss, where the sum over $\gD$ is replaced by a sum over a minibatch of data points $\gB\subset\gD$, reweighted by $\frac{|\gD|}{|\gB|}$. The MLE loss is similar, with no prior term added.

\begin{figure*}
    \centering
    \includegraphics[width=0.35\textwidth]{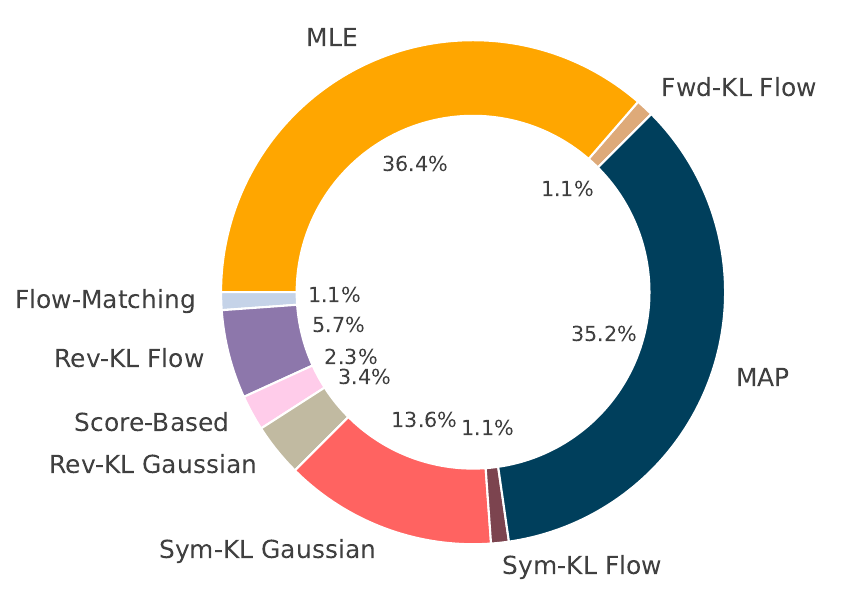}\hspace{3mm}
    \includegraphics[width=0.3\textwidth]{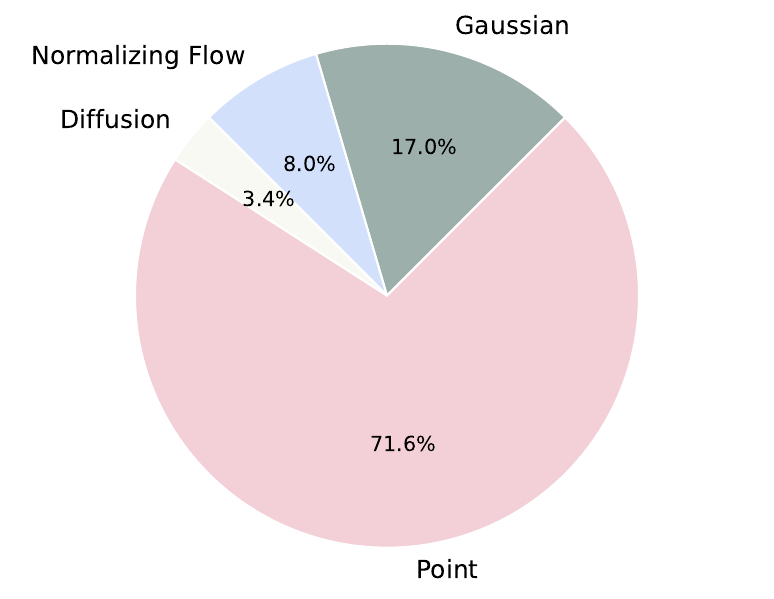}
    \includegraphics[width=0.3\textwidth]{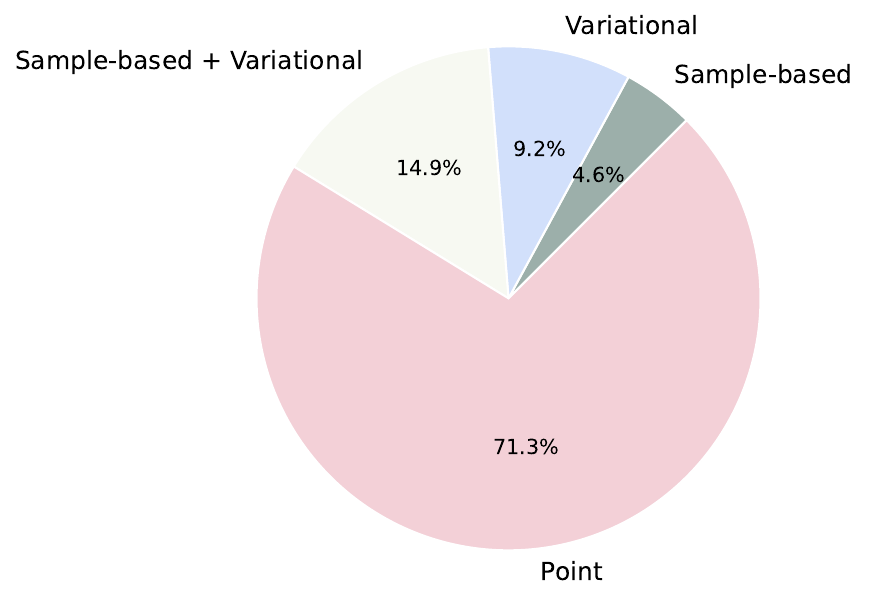}
    \vspace{-1mm}
    \caption{We plot ranking metrics aggregated over multiple tasks with varying dimensionalities, where the percentage describes the ratio of times that particular in-context estimator outperformed the others. All the different estimators compete against each other on the \textit{left} pie-chart, the \textit{middle} one describes different choices for $q_\phi$ while the \textit{right} chart describes the kind of signal used for training.}
    \vspace{-5mm}
    \label{fig:rank}
\end{figure*}

\subsection{Posterior Estimates}

For full Bayesian posterior estimation, an amortized density $q_\phi(\theta | \gD)$ approximates the true posterior $p(\theta | \gD)$. It can be trained with different objectives and distinct $q_\phi$ parametrizations, with the learning signal from either joint $(\theta, \gD)$ samples or the unnormalized target density $p(\theta | \gD) \propto p(\gD, \theta)$, or both. The details are provided below.

\subsubsection{Sample-Based Methods} 

Sample-based methods treat the approximation of $q_\phi(\theta\mid\gD)$ as a generative modeling problem. They assume that we have access to samples $(\theta,\gD)$ from the true data-generating process $\chi$. In particular, this requires the problem to be well-specified and for the generative model to expose the parameter of the conditional model, \ie, the true model proceeds via generation of $\theta$, \iid generation of inputs $\vx_i$, and generation of outputs $\vy_i$ from the model $p(\vy_i\mid \vx_i,\theta)$. 

Given samples $(\theta,\gD)\sim\chi$, we can fit a generative model, conditioned on $\gD$, to the samples $\theta$. If the objective is the log-likelihood of the samples $\theta$ under the generative model, it amounts to minimization of the forward KL divergence between the generative model and the true posterior:
\begin{align}\nonumber
    &\E_{(\theta,\gD)\sim\chi}[-\log q_\phi(\theta\mid\gD)]
    \\\label{eq:forward_kl}
    =&\E_{\gD\sim\chi}\,\KL(p(\theta\mid\gD)\|q_\phi(\theta\mid\gD))+{\rm const}.
\end{align}
Any family of generative models $q_\phi$ can be used in the approximation. However, unbiased estimates of the log-likelihood gradient in \eqref{eq:forward_kl} require that the data and ground-truth parameter values $(\theta,\gD)$ come precisely from the true data-generating process. For this to work, it is also important that the empirical distribution of $(\theta,\gD)$'s sufficiently ``covers" the region associated with the target $(\theta,\gD)$ of interest at test-time, so that $q_\phi$ will generalize well there.

\paragraph{Gaussian modelling.} One fits a Gaussian distribution, with mean and covariance output by a model (\eg, a neural network) conditioned on $\gD$. The optimal model, which minimizes \eqref{eq:forward_kl}, matches the first and second moments of the true posterior $p(\theta\mid\gD)$ for every $\gD$ in the support of $\chi$.

\paragraph{Normalizing flows}. They \citep{papamakarios2021normalizing,kobyzev2020normalizing} apply a sequence of trainable invertible transforms to convert a simple initial density, \eg~ $\gN(0, \text{I})$, to a more complex one. The invertible transformations are chosen such that the jacobian in the change of density can be easily computed, and training is done by directly optimizing the likelihood / minimizing \eqref{eq:forward_kl}.

Continuous-time normalizing flows side-step the problem of careful design of invertible transforms \citep{chen2018neural} and model $q_\phi$ as an ordinary differential equation transforming a simple distribution, \eg~ standard normal. Further, flow-matching methods~\citep{lipman2022flow,tong2023improving} provide efficient ways of training continuous-time normalizing flows in a simulation-free manner without constraining architecture choice.

\paragraph{Score-Based Diffusion}. Diffusion models~\citep{song2020score,song2020improved,song2020denoising,ho2020denoising,nichol2021improved} use stochastic differential equations to model $q_\phi$ by considering a fixed noising process and learning its reverse dynamics by estimating the time-conditioned score function. Akin to flow-matching, they are trained in a simulation-free manner and are equivalent to optimizing a variational upper bound on \eqref{eq:forward_kl}.

\begin{table*}[t]
    \centering
    \small
    % \scriptsize
    % \def\arraystretch{1.5}
    \setlength{\tabcolsep}{10pt}
    % \resizebox{\linewidth}{!}{
    \begin{tabular}{@{}l r c c c c c}
        \toprule
          & & \multicolumn{3}{c}{\textit{$L_2$ Loss} ($\downarrow$)} & \multicolumn{2}{c}{\textit{Accuracy} ($\uparrow$)}\\
         \cmidrule(lr){3-5}\cmidrule(lr){6-7}
        & \textbf{Objective} & \multicolumn{1}{c}{\textbf{GM}} & \multicolumn{1}{c}{\textbf{LR}} & \multicolumn{1}{c}{\textbf{NLR}} & \multicolumn{1}{c}{\textbf{LC}} & \multicolumn{1}{c}{\textbf{NLC}} \\
        \cmidrule(lr){3-7}
        & & $100$D & $100$D & $25$D & $100$D $2$cl & $25$D $2$cl \\
\midrule
\multirow{3}{*}{Baseline} & Random & $204.91$\sstd{$0.21$} & $105.4$\sstd{$0.3$} & $428.9$\sstd{$5.4$} & $50.4$\sstd{$0.4$} & $49.8$\sstd{$1.2$} \\
& True Posterior & $101.31$\sstd{$0.08$} & $14.5$\sstd{$0.3$} & - & - & - \\
& Optimization & $101.24$\sstd{$0.00$} & $25.1$\sstd{$0.0$} & $96.8$\sstd{$0.1$} & $70.3$\sstd{$0.0$} & $78.4$\sstd{$0.1$} \\
\cmidrule(lr){2-7}
\multirow{2}{*}{Single-Chain}
& Langevin & $102.33$\sstd{$0.03$} & $23.3$\sstd{$0.7$} & $132.3$\sstd{$1.0$} & $65.1$\sstd{$0.3$} & $73.2$\sstd{$0.3$} \\
& HMC & $102.41$\sstd{$0.03$} & $18.7$\sstd{$0.2$} & $98.1$\sstd{$0.7$} & $62.2$\sstd{$0.3$} & $70.4$\sstd{$0.1$} \\
\cmidrule(lr){2-7}
\multirow{2}{*}{Multiple-Chain} & Langevin & $101.28$\sstd{$0.00$} & $14.5$\sstd{$0.2$} & $80.2$\sstd{$0.4$} & $72.6$\sstd{$0.1$} & $79.4$\sstd{$0.2$} \\
& HMC & $101.38$\sstd{$0.00$} & $17.0$\sstd{$0.0$} & $86.4$\sstd{$0.2$} & $71.5$\sstd{$0.4$} & $76.6$\sstd{$0.1$} \\
\midrule

\multirow{3}{*}{Gaussian} & Fwd-KL &$101.38$\sstd{$0.00$} & $25.5$\sstd{$0.6$} & $276.2$\sstd{$2.2$} & $71.6$\sstd{$0.1$} & $64.9$\sstd{$0.5$} \\

& Rev-KL &$101.38$\sstd{$0.01$} & $28.5$\sstd{$0.3$} & $101.8$\sstd{$1.8$} & $72.5$\sstd{$0.1$} & \highlight{$78.5$\sstd{$0.1$}} \\

& Sym-KL &$101.37$\sstd{$0.02$} & $28.9$\sstd{$0.3$} & \highlight{$95.7$\sstd{$0.9$}} & $72.3$\sstd{$0.3$} & $78.1$\sstd{$0.2$} \\
\cmidrule(lr){2-7}

\multirow{3}{*}{Norm. Flows} & Fwd-KL &$101.39$\sstd{$0.01$} & $24.4$\sstd{$1.2$} & $268.1$\sstd{$1.9$} & $72.1$\sstd{$0.3$} & $65.4$\sstd{$0.4$} \\

& Rev-KL &$101.38$\sstd{$0.01$} & $29.1$\sstd{$1.6$} & $102.1$\sstd{$0.9$} & $72.9$\sstd{$0.1$} & \highlight{$78.6$\sstd{$0.2$}} \\

& Sym-KL &$101.37$\sstd{$0.01$} & $30.0$\sstd{$0.5$} & $103.9$\sstd{$0.4$} & $72.9$\sstd{$0.4$} & \highlight{$78.5$\sstd{$0.6$}} \\
\cmidrule(lr){2-7}

\multirow{3}{*}{Diffusion} & Score-Based &$101.42$\sstd{$0.01$} & \highlight{$22.9$\sstd{$0.3$}} & $300.4$\sstd{$2.5$} & $71.9$\sstd{$0.2$} & $64.3$\sstd{$0.1$} \\

& Flow-Matching &$101.40$\sstd{$0.02$} & $23.7$\sstd{$0.2$} & $281.6$\sstd{$1.7$} & $72.2$\sstd{$0.2$} & $65.6$\sstd{$0.4$} \\

& pDEM &$114.36$\sstd{$1.09$} & $32.7$\sstd{$0.9$} & $257.8$\sstd{$4.1$} & $72.5$\sstd{$0.2$} & $73.5$\sstd{$0.7$} \\
\cmidrule(lr){2-7}

\multirow{2}{*}{Point} & MLE &$101.30$\sstd{$0.00$} & $28.1$\sstd{$0.7$} & $99.0$\sstd{$2.9$} & $73.0$\sstd{$0.2$} & $76.5$\sstd{$0.4$} \\

& MAP & \highlight{$101.28$\sstd{$0.00$}} & $28.1$\sstd{$0.6$} & \highlight{$96.9$\sstd{$1.5$}} & \highlight{$73.4$\sstd{$0.1$}} & $78.3$\sstd{$0.2$} \\
\bottomrule
    \end{tabular}
    % }
    \vspace{-1mm}
    \caption{We compare various in-context parameter inference methods using ensemble-based predictive metrics for \textit{fixed-dimensional} estimation problems. In these experiments, each in-context learner is trained for a specific problem dimensionality. The tasks are high-dimensional, ranging from $100$ to $800$ dimensional parameters. $r$D implies $\vx \in \mathbf{R}^r$ and $s$cl implies $s$-class classification problem.}
    \vspace{-5mm}
    \label{tab:fixed_dim_ens}
\end{table*}

\subsubsection{Variational Methods}

Some methods for approximating the posterior do not rely on unbiased samples from the true data-generating process, but rather aim to fit a model to match $q(\theta\mid\gD)$ to the true posterior $p(\theta\mid\gD)$ given access to the joint $p(\theta,\gD)=p(\theta)p(\gD\mid\theta)$, to which $p(\theta\mid\gD)$ is proportional, at any $\theta$.

\paragraph{Reverse KL.} While the forward KL divergence $\KL(p(\theta\mid\gD)\|q_\phi(\theta\mid\gD))$ cannot be optimized exactly without samples from $p$, we can optimize the reverse KL divergence $\KL(q_\phi(\theta\mid\gD)\|p(\theta\mid\gD))$ exactly. The reverse KL objective can be seen as an entropy-regularized maximum likelihood:
\begin{align}\nonumber
    &\KL(q_\phi(\theta\mid\gD)\|p(\theta\mid\gD)) \\\label{eq:reverse_kl}
    =\,&\E_{\theta\sim q_\phi(\theta\mid\gD)}[-\log p(\theta\mid\gD)] - \gH[q_\phi(\theta\mid\gD)].
\end{align}
Note that there is no expectation over $\gD$, and we are free to optimize \eqref{eq:reverse_kl} for $\gD$ sampled from any distribution over datasets of interest, or even for a single fixed $\gD$. 

It is important to note that while the forward KL approaches are mean-seeking and can overestimate the variance in its approximation, the reverse KL methods are mode-seeking instead and can underestimate the variance or only model a few modes \citep{bishop2006pattern}. 

\paragraph{Diffusion samplers.} \emph{Diffusion samplers} are diffusion models fit to sample a distribution with given unnormalized density -- in this case, the joint $p(\theta)p(\gD\mid\theta)$ -- rather than to maximize a bound on log-likelihood on a data sample. Various methods for training diffusion sampler exist \citep{zhang2021path,vargas2023denoising,sendera2024improved}, most of them requiring simulation of the denoising process on each training step. One exception is the method of denoising energy matching \citep[DEM][]{akhound2024iterated}, which trains the denoiser by regressing to a biased but asymptotically constitent Monte Carlo estimate of the score function of the true posterior $p(\theta\mid\gD)$.

\begin{table*}[t]
    \centering
    \small
    % \scriptsize
    % \def\arraystretch{1.5}
    \setlength{\tabcolsep}{4 pt}
    % \resizebox{\linewidth}{!}{
    \begin{tabular}{@{}l r cc cc cc cc}
        \toprule
          & & \multicolumn{4}{c}{\textit{$L_2$ Loss} ($\downarrow$)} & \multicolumn{4}{c}{\textit{Accuracy} ($\uparrow$)}\\
         \cmidrule(lr){3-6}\cmidrule(lr){7-10}
        & \textbf{Objective} & \multicolumn{4}{c}{\textbf{NLR}} & \multicolumn{4}{c}{\textbf{NLC}} \\
        \cmidrule(lr){3-10}
        & & \multicolumn{2}{c}{\textsc{TanH}} & \multicolumn{2}{c}{\textsc{ReLU}} & \multicolumn{2}{c}{\textsc{TanH}} & \multicolumn{2}{c}{\textsc{ReLU}} \\
        \cmidrule(lr){3-10}
        & & $1$D & $25$D & $1$D & $25$D & $2$D $5$cl & $25$D $5$cl & $2$D $5$cl & $25$D $5$cl \\
\midrule
\multirow{2}{*}{Baseline} & Random & $28.1$\sstd{$0.7$} & $26.7$\sstd{$0.1$} & $533.2$\sstd{$6.1$} & $5796.0$\sstd{$93.1$} & $21.3$\sstd{$0.3$} & $19.7$\sstd{$0.3$} & $19.6$\sstd{$1.0$} & $20.4$\sstd{$0.4$} \\
& Optimization & $0.5$\sstd{$0.0$} & $25.5$\sstd{$0.0$} & $2.0$\sstd{$0.1$} & $1703.4$\sstd{$4.6$} & $88.5$\sstd{$0.1$} & $40.9$\sstd{$0.1$} & $93.8$\sstd{$0.0$} & $62.0$\sstd{$0.1$} \\
% \midrule
\cmidrule(lr){2-10}
\multirow{2}{*}{Single-Chain} & Langevin & $0.4$\sstd{$0.0$} & $28.5$\sstd{$0.1$} & \textsc{N/A} & \textsc{N/A} & $84.1$\sstd{$0.2$} & $31.4$\sstd{$0.2$} & $92.4$\sstd{$0.3$} & $52.5$\sstd{$0.3$} \\
& HMC & $0.7$\sstd{$0.0$} & $23.7$\sstd{$0.3$} & $21.7$\sstd{$1.5$} & $3905.0$\sstd{$6.8$} & $75.3$\sstd{$0.3$} & $29.6$\sstd{$0.7$} & $81.0$\sstd{$0.3$} & $52.7$\sstd{$0.3$} \\
% \midrule
\cmidrule(lr){2-10}
\multirow{2}{*}{Multiple-Chain} & Langevin & $0.3$\sstd{$0.0$} & $15.8$\sstd{$0.0$} & \textsc{N/A} & \textsc{N/A} & $87.9$\sstd{$0.1$} & $41.2$\sstd{$0.3$} & $94.0$\sstd{$0.1$} & $63.6$\sstd{$0.3$} \\
& HMC & $0.7$\sstd{$0.0$} & $17.3$\sstd{$0.0$} & $20.3$\sstd{$0.2$} & $3825.6$\sstd{$1.5$} & $77.3$\sstd{$0.2$} & $40.3$\sstd{$0.2$} & $83.3$\sstd{$0.1$} & $60.8$\sstd{$0.2$} \\
\midrule
\multirow{3}{*}{Gaussian} & Fwd-KL &$28.1$\sstd{$0.7$} & $26.7$\sstd{$0.1$} & $277.7$\sstd{$5.6$} & $3657.1$\sstd{$34.5$} & $22.0$\sstd{$0.4$} & $20.3$\sstd{$0.6$} & $51.6$\sstd{$1.3$} & $45.5$\sstd{$0.5$} \\
& Rev-KL &$0.6$\sstd{$0.0$} & \highlight{$23.0$\sstd{$3.9$}} & \highlight{$2.0$\sstd{$0.1$}} & \highlight{$1831.7$\sstd{$105.2$}} & $64.7$\sstd{$1.0$} & $19.6$\sstd{$0.5$} & $76.6$\sstd{$4.3$} & $27.3$\sstd{$0.4$} \\
& Sym-KL &$1.4$\sstd{$0.0$} & $25.8$\sstd{$0.0$} & $3.6$\sstd{$0.9$} & \highlight{$1735.6$\sstd{$23.3$}} & $21.6$\sstd{$0.4$} & $19.9$\sstd{$0.5$} & $62.5$\sstd{$0.5$} & $26.6$\sstd{$0.8$} \\
\cmidrule(lr){2-10}
\multirow{3}{*}{Norm. Flows} & Fwd-KL &$27.9$\sstd{$0.6$} & $26.9$\sstd{$0.1$} & $239.8$\sstd{$16.5$} & $3330.0$\sstd{$37.9$} & $20.9$\sstd{$1.2$} & $20.3$\sstd{$0.4$} & $55.1$\sstd{$0.9$} & $47.5$\sstd{$0.3$} \\
& Rev-KL &$0.5$\sstd{$0.0$} & $25.8$\sstd{$0.0$} & \highlight{$2.2$\sstd{$0.3$}} & $1823.7$\sstd{$50.7$} & $28.2$\sstd{$19.0$} & $20.0$\sstd{$0.5$} & $79.5$\sstd{$0.8$} & $52.7$\sstd{$0.5$} \\
& Sym-KL &$0.5$\sstd{$0.1$} & $25.8$\sstd{$0.0$} & \highlight{$3.0$\sstd{$1.6$}} & $1810.8$\sstd{$71.1$} & $19.5$\sstd{$1.4$} & $20.0$\sstd{$0.5$} & $79.7$\sstd{$0.7$} & $53.0$\sstd{$0.2$} \\
\cmidrule(lr){2-10}
\multirow{3}{*}{Diffusion} & Score-Based &$29.6$\sstd{$0.9$} & $28.1$\sstd{$0.3$} & $361.9$\sstd{$20.9$} & $4684.6$\sstd{$127.5$} & $20.1$\sstd{$0.5$} & $19.7$\sstd{$0.4$} & $35.2$\sstd{$1.0$} & $35.8$\sstd{$1.0$} \\
& Flow-Matching &$28.4$\sstd{$1.0$} & $27.0$\sstd{$0.2$} & $271.7$\sstd{$11.9$} & $3646.3$\sstd{$69.9$} & $20.2$\sstd{$1.8$} & $20.0$\sstd{$0.5$} & $50.0$\sstd{$1.6$} & $45.0$\sstd{$0.8$} \\
& pDEM &$29.5$\sstd{$0.8$} & $26.0$\sstd{$0.1$} & $770.8$\sstd{$684.5$} & $4179.7$\sstd{$177.3$} & $20.4$\sstd{$0.9$} & $19.7$\sstd{$0.7$} & $60.9$\sstd{$0.5$} & $51.9$\sstd{$0.5$} \\
\cmidrule(lr){2-10}
\multirow{2}{*}{Point} & MLE &$0.4$\sstd{$0.0$} & \highlight{$21.0$\sstd{$0.2$}} & $3.0$\sstd{$0.3$} & $1899.2$\sstd{$38.8$} & \highlight{$89.0$\sstd{$0.1$}} & \highlight{$40.2$\sstd{$0.7$}} & \highlight{$93.9$\sstd{$0.1$}} & $60.8$\sstd{$0.5$} \\
& MAP & \highlight{$0.3$\sstd{$0.0$}} & \highlight{$20.8$\sstd{$0.3$}} & $2.8$\sstd{$0.3$} & $1919.2$\sstd{$52.1$} & $86.0$\sstd{$0.1$} & $20.0$\sstd{$0.4$} & $93.1$\sstd{$0.1$} & \highlight{$61.7$\sstd{$0.2$}} \\
\bottomrule
    \end{tabular}
    % }
    \vspace{-1mm}
    \caption{Comparison of various in-context estimators in inferring the parameters of a $2$-layered neural network for nonlinear regression and classification tasks of varying dimensionalities, number of classes and activation functions. Amortized point estimators considerably outperform posterior counterparts, especially for high-dimensional classification tasks.}
    \vspace{-5mm}
    \label{tab:fixed_dim_2layer_ens}
\end{table*}
\paragraph{MCMC.} While they are not amortized variational methods, Monte Carlo Markov chain (MCMC)~\citep{welling2011bayesian,hoffman2014no,chen2014stochastic} methods can be used to draw samples from $p(\theta\mid\gD)$ given access only to the joint $p(\theta,\gD)=p(\theta)p(\gD\mid\theta)$. The samples can then be used to estimate the expectation defining the posterior predictive \eqref{eq:postpred}.
In general, MCMC methods have a guarantee of convergence to the target distribution given enough samples or iterations, but convergence can be slow \citep{andrieu2003introduction,gilks1996strategies,neal2012mcmc}. Some MCMC methods, such as Langevin and Hamiltonian MCMC, also require access to the gradient of the log-likelihood.

\subsubsection{Sample-based + Variational Methods} 
One can combine the two estimation procedures outlined above. We consider an equally-weighted combination of forward and reverse KL as the divergence metric, called symmetric KL, for learning $q_\phi$ in cases where it is modeled as a Gaussian distribution or a discrete normalizing flow.

\begin{table*}[t]
    \centering
    \small
    % \scriptsize
    % \def\arraystretch{1.5}
    \setlength{\tabcolsep}{10pt}
    \begin{tabular}{@{}l r c c c c c}
        \toprule
          & & \multicolumn{3}{c}{\textit{$L_2$ Loss} ($\downarrow$)} & \multicolumn{2}{c}{\textit{Accuracy} ($\uparrow$)}\\
         \cmidrule(lr){3-5}\cmidrule(lr){6-7}
        & \textbf{Objective} & \multicolumn{1}{c}{\textbf{GM}} & \multicolumn{1}{c}{\textbf{LR}} & \multicolumn{1}{c}{\textbf{NLR}} & \multicolumn{1}{c}{\textbf{LC}} & \multicolumn{1}{c}{\textbf{NLC}} \\
        \cmidrule(lr){3-7}
        & & $100$D & $100$D & $50$D & $100$D $2$cl & $50$D $2$cl \\
\midrule
\multirow{2}{*}{Baseline} & Random & $202.25$\sstd{$0.41$} & $103.4$\sstd{$0.7$} & $914.7$\sstd{$12.1$} & $50.3$\sstd{$0.4$} & $49.5$\sstd{$1.3$} \\
& Optimization & $100.88$\sstd{$0.00$} & $20.1$\sstd{$0.0$} & $301.2$\sstd{$0.1$} & $71.3$\sstd{$0.0$} & $76.5$\sstd{$0.0$} \\
% \midrule
\cmidrule(lr){2-7}
\multirow{2}{*}{Single-Chain} & Langevin & $101.91$\sstd{$0.03$} & $21.4$\sstd{$0.8$} & \textsc{N/A} & $65.4$\sstd{$0.4$} & $69.9$\sstd{$0.4$} \\
& HMC & $102.02$\sstd{$0.02$} & $17.7$\sstd{$0.1$} & $303.3$\sstd{$2.4$} & $62.7$\sstd{$0.2$} & $68.2$\sstd{$0.4$} \\
% \midrule
\cmidrule(lr){2-7}
\multirow{2}{*}{Multiple-Chain} &
Langevin & $100.91$\sstd{$0.00$} & $13.2$\sstd{$0.1$} & \textsc{N/A} & $72.9$\sstd{$0.2$} & $76.8$\sstd{$0.1$} \\
& HMC & $100.99$\sstd{$0.01$} & $15.9$\sstd{$0.0$} & $285.6$\sstd{$0.3$} & $71.7$\sstd{$0.2$} & $75.3$\sstd{$0.3$} \\
\midrule

\multirow{3}{*}{Gaussian} & Fwd-KL &$103.96$\sstd{$0.07$} & $33.4$\sstd{$1.0$} & $564.8$\sstd{$5.7$} & $71.0$\sstd{$0.3$} & $66.9$\sstd{$0.4$} \\

& Rev-KL & \highlight{$102.66$\sstd{$0.07$}} & $31.3$\sstd{$0.6$} & $278.9$\sstd{$2.5$} & \highlight{$72.0$\sstd{$0.3$}} & \highlight{$76.9$\sstd{$0.2$}} \\

& Sym-KL &$102.67$\sstd{$0.06$} & $30.9$\sstd{$1.0$} & \highlight{$267.9$\sstd{$0.8$}} & \highlight{$72.1$\sstd{$0.4$}} & $76.6$\sstd{$0.3$} \\
% \midrule
\cmidrule(lr){2-7}

\multirow{3}{*}{Norm. Flows} & Fwd-KL &$103.71$\sstd{$0.10$} & $32.3$\sstd{$0.6$} & $551.5$\sstd{$3.9$} & $71.2$\sstd{$0.2$} & $67.5$\sstd{$0.3$} \\

& Rev-KL &$102.76$\sstd{$0.06$} & $32.0$\sstd{$0.6$} & $274.4$\sstd{$2.2$} & $71.9$\sstd{$0.2$} & \highlight{$77.0$\sstd{$0.2$}} \\

& Sym-KL &$102.72$\sstd{$0.06$} & $44.5$\sstd{$6.0$} & $274.3$\sstd{$4.5$} & $71.9$\sstd{$0.2$} & $70.3$\sstd{$1.0$} \\
% \midrule
\cmidrule(lr){2-7}

\multirow{3}{*}{Diffusion} & Score-Based &$103.20$\sstd{$0.07$} & \highlight{$28.0$\sstd{$0.7$}} & $671.7$\sstd{$13.7$} & $70.8$\sstd{$0.5$} & $66.5$\sstd{$0.4$} \\

& Flow-Matching &$102.84$\sstd{$0.04$} & \highlight{$27.9$\sstd{$0.7$}} & $579.4$\sstd{$2.1$} & $71.0$\sstd{$0.6$} & $67.6$\sstd{$0.2$} \\

& pDEM &$114.61$\sstd{$0.71$} & $57.8$\sstd{$12.4$} & $560.0$\sstd{$7.7$} & $71.6$\sstd{$0.3$} & $68.3$\sstd{$0.2$} \\
% \midrule
\cmidrule(lr){2-7}

\multirow{2}{*}{Point} & MLE &$103.07$\sstd{$0.20$} & $31.4$\sstd{$0.4$} & $289.1$\sstd{$3.2$} & $70.9$\sstd{$0.2$} & $75.5$\sstd{$0.2$} \\

& MAP & \highlight{$102.60$\sstd{$0.07$}} & $31.2$\sstd{$0.5$} & $285.7$\sstd{$2.0$} & \highlight{$72.3$\sstd{$0.2$}} & $76.3$\sstd{$0.2$} \\
\bottomrule
    \end{tabular}
    \vspace{-1mm}
    \caption{Experiments on \textit{variable dimensional} problems evaluate in-context parameter estimators where a single in-context model is trained for each column, and can jointly perform any of the lower dimensional tasks in the same family. Analysis is done through the ensemble-based prediction metrics, with further results on lower dimensional task counterparts in \cref{apdx:variable-dim}.}
    \vspace{-5mm}
    \label{tab:variable_dim_ens}
\end{table*}
\section{Experiments}

Our goal in this comparative study is to evaluate the amortized estimation procedures discussed in \cref{sec:training_objectives} for both in-distribution (ID) and out-of-distribution (OoD) generalization. We consider variants of point-estimation methods, forward and reverese KL approaches including diffusion and normalizing flows, and symmetric KL objective, with a focus on explicit conditioning on the set of observations.
We evaluate the Bayesian and frequentist in-context estimators on a wide suite of probabilistic models through the lens of predictive performance, and discuss the suite of tasks, baselines and metrics considered in this study below.

\looseness=-1
\textbf{Tasks}.
We consider estimating the mean of a Gaussian distribution (GM), means of a Gaussian Mixture Model (GMM), parameters of (non-)linear regression (NLR/LR) and classification (NLC/LC) models, where the nonlinear problems are modeled through a neural network and the parameters correspond to the parameters of the network. We refer the readers to \cref{apdx:probabilistic_models} for details about the probabilistic models, including the likelihood and prior considered in each setup. We further evaluate the different in-context estimators on OoD transfer in the case of model misspecification and its applications to real-world tabular tasks, where the data-generating distribution shifts between training and evaluation.

\textbf{Baselines}. To understand whether the in-context estimators achieve reasonable performance, we consider multiple non-amortized procedures as reference: sampling from the prior (Random), the true posterior (True Posterior), iterative sampling procedures like single or multiple chains of Langevin and Hamiltonian (HMC) MCMC, and optimizing the parameters through MLE (Optimization). 

\textbf{Metrics}. Aligned with our goal towards better predictions, we leverage predictive metrics like $L_2$ loss and accuracy under the parameters inferred by the in-context estimators. This provides us two choices of metrics: expected loss/accuracy and ensemble based metric. For regression problems, the former can be seen as
\begin{align}
    \mathbb{E}_{\vx_*,\vy_*, \gD \sim \chi}\mathbb{E}_{\theta \sim q_\phi(\cdot | \gD)} \norm{\hat{\vy} - \vy}^2
\end{align}
where $\hat{\vy}$ is the mode of $p(\vy | \vx, \theta)$, while the ensembling based metric can be seen as
\begin{align}
    \mathbb{E}_{\vx_*,\vy_*, \gD \sim \chi} \norm{\mathbb{E}_{\theta \sim q_\phi(\cdot | \gD)}\hat{\vy} - \vy}^2
\end{align}
Similarly for classification, we rely on accuracy instead of $L_2$ loss and consider the mode of different $\hat{\vy}$ for ensembling as opposed to averaging.

We primarily consider the ensemble-based metric for evaluation since it is easily available for full posterior approximations due to the ease of sampling from them. Additionally, since point estimation methods only provide a single parameter value, the two metrics are trivially the same in their case. See \cref{apdx:metrics} for details on the metrics.

\begin{table}[t]
    \centering
    % \small
    % \def\arraystretch{1.05}
    \resizebox{\linewidth}{!}{
    \setlength{\tabcolsep}{1pt}
    \begin{tabular}{@{}llccc}
    \toprule
     \multirow{2}{*}{$\chi_{train}\;(\rightarrow)$} & \multirow{2}{*}{\textit{Data}} & \multirow{2}{*}{Linear} & MLP & GP \\
     & & & Nonlinear & Nonlinear \\
    \cmidrule{2-5}
    $\chi_{test}\;(\rightarrow)$&  \textit{Model} & NLR & LR & NLR \\
    \midrule
\multirow{4}{*}{Gaussian} & Fwd-KL &$2.618$\sstd{$0.176$} & $2.042$\sstd{$0.122$} & $1.216$\sstd{$0.078$} \\
& Sym-KL &$0.371$\sstd{$0.007$} & $1.696$\sstd{$0.087$} & $0.206$\sstd{$0.009$} \\
& Rev-KL &$0.284$\sstd{$0.002$} & $1.695$\sstd{$0.124$} & $0.052$\sstd{$0.001$} \\
& \quad \textit{+ switched data} &$0.277$\sstd{$0.002$} & \highlight{$1.221$\sstd{$0.003$}} & $0.029$\sstd{$0.003$} \\
\midrule

\multirow{4}{*}{Norm. Flows} & Fwd-KL &$0.657$\sstd{$0.041$} & $1.691$\sstd{$0.129$} & $0.563$\sstd{$0.057$} \\
& Sym-KL &$0.277$\sstd{$0.002$} & $1.403$\sstd{$0.049$} & $0.043$\sstd{$0.003$} \\
& Rev-KL &$0.276$\sstd{$0.002$} & $1.445$\sstd{$0.029$} & $0.045$\sstd{$0.004$} \\
& \quad \textit{+ switched data} &$0.269$\sstd{$0.002$} & \highlight{$1.220$\sstd{$0.005$}} & $0.026$\sstd{$0.003$} \\
\midrule

\multirow{4}{*}{Diffusion} & Score-Based &$0.459$\sstd{$0.025$} & $1.428$\sstd{$0.090$} & $0.328$\sstd{$0.012$} \\
& Flow-Matching &$0.797$\sstd{$0.083$} & $1.806$\sstd{$0.085$} & $0.541$\sstd{$0.063$} \\
& pDEM &$0.848$\sstd{$0.115$} & $1.727$\sstd{$0.169$} & $0.617$\sstd{$0.062$} \\
& \quad \textit{+ switched data} &$0.792$\sstd{$0.106$} & $1.682$\sstd{$0.700$} & $0.889$\sstd{$0.116$} \\
\midrule

\multirow{4}{*}{Point} & MLE &$0.399$\sstd{$0.019$} & $1.525$\sstd{$0.046$} & $0.027$\sstd{$0.016$} \\
& \quad \textit{+ switched data} &$0.382$\sstd{$0.007$} & $1.227$\sstd{$0.003$} & \highlight{$0.002$\sstd{$0.000$}} \\
& MAP &$0.267$\sstd{$0.001$} & $1.541$\sstd{$0.081$} & $0.025$\sstd{$0.000$} \\
& \quad \textit{+ switched data} & \highlight{$0.263$\sstd{$0.000$}} & $1.225$\sstd{$0.004$} & $0.014$\sstd{$0.000$} \\
\bottomrule
    \end{tabular}
    }
    \vspace{-2mm}
    \caption{Evaluating OoD generalization to novel tasks with the mapping from $\vx$ to $\vy$ altered from training. $\chi_{train}$ is the data-generating process under the assumed model $p$ (prior and likelihood) while $\chi_{test}$ is the data that we want to generalize to. By default, $\gD \sim \chi_{train}$ is used for training except in the case of switched data, where point and variational approaches are trained on data different from the assumed model. Evaluation is done through the ensembled predictive $L_2$ loss. (N-)LR represents (non-)linear regression, and GP represents Gaussian Process.}
    \vspace{-6mm}
    \label{tab:misspecification_ens}
\end{table}

\subsection{Evaluating in-distribution parameter inference}

In-context estimators, whether point or posterior, can be leveraged to generalize to novel tasks zero-shot after being trained over multiple different datasets $\gD_{\rm train} \sim \chi_{\rm train}$. We first test for in-distribution generalization by sampling novel tasks $\gD_{\rm test} \sim \chi_{\rm train}$ and evaluating how well parameter samples generalize under the predictive metrics on $\gD_{\rm test}$. The benchmark consists of $88$ tasks, where each task is defined by a different probabilistic model configuration, leading to the training of $324$ models for each in-context estimator considered.

We provide a high-level visualization of the outcome of our experiments in \cref{fig:rank}, which demonstrates the proportion of tasks each (class of) estimator outperformed its counterparts. This aggregation is based on a winner-take-all ranking procedure across all the tasks, where the performance of each estimator for each task is averaged over $6$ seeds.

Our experiments indicate that in-context point estimation procedures outperform Bayesian methods, with a roughly even split between the amortized MLE and MAP estimator. This points to the inability of amortized Bayesian estimators in modeling the posterior well, as it was always maintained that the underlying modeling assumption (\ie the parametric form of the likelihood) as well as the prior considered matched the true data-generating distribution $\chi$ in these tasks. Within Bayesian methods, Gaussian assumption outperformed more sophisticated methods like normalizing flows and diffusion models with the symmetric KL divergence training procedure (Sample + Variational) being the dominant approach to posterior estimation.

\subsubsection{Fixed-Dimensional}

In-context learning relies on a common model to solve novel tasks and is thus limited in generalization to cases where the input and output spaces are shared across tasks, for, \eg, scalar inputs and outputs for $1$-dimensional regression or a fixed vocabulary in LLMs. Similarly, parametric inference predicts, or describes a distribution over, $\theta$ and consequently requires its size to be shared across different problems. This is inherently defined by the probabilistic model, \ie the likelihood and prior implicitly define the space of parameters. Thus, naive training of in-context estimators requires a different model to be trained for a $1$-dimensional problem than that for a $2$-dimensional one. 

\looseness=-1
We evaluate the in-context estimators on the suite of probabilistic models in \cref{tab:fixed_dim_ens}, using the ensemble-based predictive performance as the metrics. 
Our experiments on high-dimensional versions of the respective probabilistic models demonstrate that point estimates are competitive and often better than Bayesian counterparts, even without the benefit of ensembling. In these experiments, for nonlinear models we consider a single layered neural network with \textsc{ReLU} activation function. To test the estimators on even higher dimensional problems, we next consider a $2$-layered neural network with \textsc{TanH} or \textsc{ReLU} activation in \cref{tab:fixed_dim_2layer_ens} which highlights clear superiority of amortized point estimators over posterior counterparts. See \cref{apdx:fixed-dim} for details.

\begin{table}[t]
    \centering
    \small
    \setlength{\tabcolsep}{1pt}
    \resizebox{\linewidth}{!}{
    \begin{tabular}{@{}lc cccc}
        \toprule
         &  & \multicolumn{2}{c}{\textit{$L_2$ Loss} ($\downarrow)$} & \multicolumn{2}{c}{\textit{Accuracy} ($\uparrow$)}\\
        \cmidrule(lr){3-4}\cmidrule(lr){5-6}
        & \textbf{Model} & LR & NLR & LC & NLC \\
        \midrule
Random & - & $11.77$\sstd{$0.21$} & $18.04$\sstd{$0.98$} & $49.87$\sstd{$1.28$} & $53.37$\sstd{$3.19$} \\
\midrule
Fwd-KL & \multirow{3}{*}{\rotatebox[origin=c]{90}{Gaussian}} &$7.91$\sstd{$0.90$} & $17.18$\sstd{$1.51$} & $72.70$\sstd{$5.56$} & $71.40$\sstd{$1.71$} \\
Rev-KL & &$7.68$\sstd{$0.68$} & $7.80$\sstd{$1.06$} & $76.65$\sstd{$5.75$} & \highlight{$77.04$\sstd{$4.91$}} \\
Sym-KL & &$7.39$\sstd{$0.55$} & $24.65$\sstd{$18.02$} & $75.97$\sstd{$2.68$} & $78.46$\sstd{$2.90$} \\
\midrule
Fwd-KL & \multirow{3}{*}{\rotatebox[origin=c]{90}{Flow}} &$8.07$\sstd{$0.35$} & $14.29$\sstd{$0.74$} & $72.90$\sstd{$2.85$} & $71.41$\sstd{$2.31$} \\
Rev-KL & &$7.57$\sstd{$0.49$} & \highlight{$8.48$\sstd{$1.69$}} & $74.91$\sstd{$5.74$} & \highlight{$81.03$\sstd{$2.61$}} \\
Sym-KL & &$7.48$\sstd{$0.44$} & \highlight{$9.76$\sstd{$4.68$}} & $79.23$\sstd{$2.75$} & \highlight{$81.37$\sstd{$1.55$}} \\
\midrule
Score-Based & \multirow{3}{*}{\rotatebox[origin=c]{90}{Diffusion}} &$8.63$\sstd{$3.93$} & $31.95$\sstd{$2.96$} & $61.73$\sstd{$10.04$} & $70.36$\sstd{$1.24$} \\
Flow-Matching & &$8.77$\sstd{$2.71$} & $18.21$\sstd{$2.00$} & $75.72$\sstd{$3.10$} & $72.91$\sstd{$0.65$} \\
pDEM & & \highlight{$6.09$\sstd{$0.23$}} & $12.36$\sstd{$0.41$} & \highlight{$82.90$\sstd{$1.07$}} & $74.88$\sstd{$0.19$} \\
\midrule
MLE & \multirow{2}{*}{\rotatebox[origin=c]{90}{Point}} &$7.19$\sstd{$0.27$} & \highlight{$7.88$\sstd{$1.92$}} & $76.19$\sstd{$3.54$} & \highlight{$79.05$\sstd{$3.91$}} \\
MAP & &$7.18$\sstd{$0.36$} & \highlight{$7.71$\sstd{$1.31$}} & $79.82$\sstd{$1.10$} & \highlight{$78.75$\sstd{$3.32$}} \\
\bottomrule
    \end{tabular}
    }
    \vspace{-2mm}
    \caption{We provide a comparative analysis between various estimators on OoD generalization to novel real-world tasks from the OpenML platform after being trained solely on simulated data, evaluated through ensembled predictive loss and accuracy metrics.}
    \vspace{-6mm}
    \label{tab:tabular_ens}
\end{table}

\subsubsection{Variable-Dimensional}

\label{sec:variable-dim}
Next, we alleviate the limitation of fixed-dimensional parametric inference by embedding lower-dimensional problems into a fixed higher dimension. For example, a $1$-dimensional linear regression model can be embedded in $100$-dimensional space with the additional parameters set to $0$. This simple masking procedure allows the in-context estimators to generalize to tasks with variable number of features, leading to the same estimator solving problems with different number of features. We evaluate the estimators on the variable-dimensional setup in \cref{tab:variable_dim_ens} and again see that amortized point estimation methods remain competitive and often better than Bayesian counterparts. Refer to \ref{apdx:variable-dim} for additional experiments and details.

\subsection{Misspecification}

Having studied in-distribution generalization, we now turn to cases of OoD generalization where the evaluation datasets are sampled from a different, sometimes unknown, distribution $\gD_{\rm test} \sim \chi_{\rm test}$. We study two cases: controlled synthetic and real-world tabular problems. This analysis is aimed to test the estimators' ability to handle changes in the underlying ground-truth mapping $p(y | \vx)$ as well as when $\vx$ follows a different distribution at evaluation, which is important since we often do not know the underlying model that generates the data of interest.

\subsubsection{Synthetic}

\looseness=-1
We consider $1$-dimensional regression problems with different underlying mappings $p(\vy | \vx)$ between training and evaluation. Here, we are interested in generalizing to data obtained from $\chi_{\rm test}$ but we assume that we do not know the underlying parametric form for this data. Instead, we assume a parametric form $p(\vy | \vx, \theta)$ which leads to $\chi_{\rm train}$ as the data-generating distribution, with $\chi_{\rm train} \neq \chi_{\rm test}$. 

Given this misspecification, sample-based in-context estimators can only be trained on $\gD_{\rm train} \sim \chi_{\rm train}$, however point estimation procedures and variational methods don't have this limitation, and can be trained with $\gD_{\rm train} \sim \chi_{\rm test}$ if sampling from $\chi_{\rm test}$ is relatively easy.

\cref{tab:misspecification_ens} highlights the performance of different estimators when evaluated on $\chi_{\rm test}$ and trained on $\chi_{\rm train}$, except for ``+ switched data" where even during training datasets are sampled from $\chi_{\rm test}$. We see that point estimators and variational methods can lead to better predictions by being trained directly on $\chi_{\rm test}$, with point estimators outperforming others in general. We refer to \cref{apdx:misspecification} for details.

\subsubsection{Tabular}

Finally, we turn our attention to a suite of regression and classification tasks from the OpenML platform, filtered from \textit{OpenML-CTR23 - A curated tabular regression benchmarking suite}~\citep{fischer2023openmlctr23} and \textit{OpenML-CC18 Curated Classification benchmark
}~\citep{bischl2019openmlcc18}. We exclude tasks with missing values, or with more than $100$ features. This presents a case of extreme OoD generalization as we use the in-context estimators trained in \cref{sec:variable-dim} and test their generalization zero-shot through inference of (non-)linear regression and classification assumptions on a suite of $9$ and $13$ tabular problems, respectively.

We see in \cref{tab:tabular_ens} that point estimation procedures and variational methods perform better than those trained based on samples, where we consider an average over $6$ seeds and use a $5$-fold cross validation to obtain train / test splits for each dataset. We refer to \ref{apdx:tabular} for details. 

\section{Conclusion}

Our simulations in the amortized setting suggest that point estimation methods tend to outperform distribution estimators for posterior predictive modeling, especially on problems where the posterior over parameters is high-dimensional and multimodal. While one potential reason for this is the suboptimality or sample-inefficiency of the training objectives and model architectures, which research on amortized inference should continue to improve, our findings may be indicative of a more fundamental obstacle in Bayesian modeling. Many multimodal problems exhibit a large number of distinct modes that lead to equivalent solutions, but still require increasing expressivity in the approximate posterior to represent each of these, \emph{potentially redundant}, modes. The latter challenge is manifested in more complex problems as well, \eg, the identifiability problem in mixture models \citep{teicher1963identifiability,yakowitz1968identifiability} and symmetries in Bayesian neural networks, where the number of modes has a combinatorial explosion in the network width, but mode connectivity results show that the posterior does not have high energy barriers \citep{draxler2019essentially}, especially modulo symmetries \citep{ferbach2024proving}. 

While those results concern the non-amortized setting, we have shown that when in-context parameter estimation is considered, the same challenges arise even in simple models. This points to the need for hybrid approaches to amortized inference, drawing from non-amortized methods where only a subset of parameters undergo a Bayesian treatment \citep{daxberger2022bayesian} and amortized variational families are chosen to represent posteriors more efficiently \citep{sharma2023bayesian,doan2025bayesian}.

\section*{Acknowledgements}
The authors would like to acknowledge the computing resources provided by the Mila cluster to enable the experiments outlined in this work. SM acknowledges the support of UNIQUE's scholarship.
GL acknowledges support from the Canada CIFAR AI Chair program, and the Canada Research Chair in Neural Computations and Interfacing. YB acknowledges the support from CIFAR and the CIFAR AI Chair program as well as NSERC funding for the Herzberg Canada Gold medal.
The authors also thank NVIDIA for computing resources.

\section*{Impact Statement}
We evaluate different in-context estimators for the task of inferring parameters for better downstream predictions. The goal of this work is to provide a rigorous and comparative study for the advancement of the general field of machine learning. There are many potential societal
consequences of our work, none which we feel must be
specifically highlighted here.
\clearpage
\bibliography{bibliography}

\begin{thebibliography}{64}
\providecommand{\natexlab}[1]{#1}
\providecommand{\url}[1]{\texttt{#1}}
\expandafter\ifx\csname urlstyle\endcsname\relax
  \providecommand{\doi}[1]{doi: #1}\else
  \providecommand{\doi}{doi: \begingroup \urlstyle{rm}\Url}\fi

\bibitem[Adlam et~al.(2020)Adlam, Snoek, and Smith]{adlam2020cold}
Adlam, B., Snoek, J., and Smith, S.~L.
\newblock Cold posteriors and aleatoric uncertainty.
\newblock \emph{arXiv preprint arXiv:2008.00029}, 2020.

\bibitem[Akhound-Sadegh et~al.(2024)Akhound-Sadegh, Rector-Brooks, Bose, Mittal, Lemos, Liu, Sendera, Ravanbakhsh, Gidel, Bengio, et~al.]{akhound2024iterated}
Akhound-Sadegh, T., Rector-Brooks, J., Bose, A.~J., Mittal, S., Lemos, P., Liu, C.-H., Sendera, M., Ravanbakhsh, S., Gidel, G., Bengio, Y., et~al.
\newblock Iterated denoising energy matching for sampling from boltzmann densities.
\newblock \emph{International Conference on Machine Learning (ICML)}, 2024.

\bibitem[Aky{\"u}rek et~al.(2023)Aky{\"u}rek, Schuurmans, Andreas, Ma, and Zhou]{akyurek2022learning}
Aky{\"u}rek, E., Schuurmans, D., Andreas, J., Ma, T., and Zhou, D.
\newblock What learning algorithm is in-context learning? investigations with linear models.
\newblock \emph{International Conference on Learning Representations (ICLR)}, 2023.

\bibitem[Albergo et~al.(2023)Albergo, Boffi, and Vanden-Eijnden]{albergo2023stochastic}
Albergo, M.~S., Boffi, N.~M., and Vanden-Eijnden, E.
\newblock Stochastic interpolants: A unifying framework for flows and diffusions.
\newblock \emph{arXiv preprint arXiv:2303.08797}, 2023.

\bibitem[Albergo et~al.(2024)Albergo, Goldstein, Boffi, Ranganath, and Vanden-Eijnden]{albergo2023dependent}
Albergo, M.~S., Goldstein, M., Boffi, N.~M., Ranganath, R., and Vanden-Eijnden, E.
\newblock Stochastic interpolants with data-dependent couplings.
\newblock \emph{International Conference on Machine Learning (ICML)}, 2024.

\bibitem[Anderson(1982)]{anderson1982reverse}
Anderson, B.~D.
\newblock Reverse-time diffusion equation models.
\newblock \emph{Stochastic Processes and their Applications}, 12\penalty0 (3):\penalty0 313--326, 1982.

\bibitem[Andrieu et~al.(2003)Andrieu, De~Freitas, Doucet, and Jordan]{andrieu2003introduction}
Andrieu, C., De~Freitas, N., Doucet, A., and Jordan, M.~I.
\newblock An introduction to {MCMC} for machine learning.
\newblock \emph{Machine learning}, 50:\penalty0 5--43, 2003.

\bibitem[Bischl et~al.(2021)Bischl, Casalicchio, Feurer, Hutter, Lang, Mantovani, van Rijn, and Vanschoren]{bischl2019openmlcc18}
Bischl, B., Casalicchio, G., Feurer, M., Hutter, F., Lang, M., Mantovani, R.~G., van Rijn, J.~N., and Vanschoren, J.
\newblock Openml benchmarking suites.
\newblock \emph{Neural Information Processing Systems (NeurIPS) Datasets and Benchmarks Track}, 2021.

\bibitem[Bishop \& Nasrabadi(2006)Bishop and Nasrabadi]{bishop2006pattern}
Bishop, C.~M. and Nasrabadi, N.~M.
\newblock \emph{Pattern recognition and machine learning}, volume~4.
\newblock Springer, 2006.

\bibitem[Blei et~al.(2017)Blei, Kucukelbir, and McAuliffe]{blei2017variational}
Blei, D.~M., Kucukelbir, A., and McAuliffe, J.~D.
\newblock Variational inference: A review for statisticians.
\newblock \emph{Journal of the American statistical Association}, 112\penalty0 (518):\penalty0 859--877, 2017.

\bibitem[Blessing et~al.(2024)Blessing, Jia, Esslinger, Vargas, and Neumann]{blessing2024beyond}
Blessing, D., Jia, X., Esslinger, J., Vargas, F., and Neumann, G.
\newblock Beyond {ELBOs}: A large-scale evaluation of variational methods for sampling.
\newblock \emph{International Conference on Machine Learning (ICML)}, 2024.

\bibitem[Chen et~al.(2018)Chen, Rubanova, Bettencourt, and Duvenaud]{chen2018neural}
Chen, R.~T., Rubanova, Y., Bettencourt, J., and Duvenaud, D.~K.
\newblock Neural ordinary differential equations.
\newblock \emph{Neural Information Processing Systems (NIPS)}, 2018.

\bibitem[Chen et~al.(2014)Chen, Fox, and Guestrin]{chen2014stochastic}
Chen, T., Fox, E., and Guestrin, C.
\newblock Stochastic gradient {Hamiltonian Monte Carlo}.
\newblock \emph{International Conference on Machine Learning (ICML)}, 2014.

\bibitem[Cremer et~al.(2018)Cremer, Li, and Duvenaud]{cremer2018inference}
Cremer, C., Li, X., and Duvenaud, D.
\newblock Inference suboptimality in variational autoencoders.
\newblock \emph{International Conference on Machine Learning (ICML)}, 2018.

\bibitem[Daxberger et~al.(2021)Daxberger, Nalisnick, Allingham, Antorán, and Hernández-Lobato]{daxberger2022bayesian}
Daxberger, E., Nalisnick, E., Allingham, J.~U., Antorán, J., and Hernández-Lobato, J.~M.
\newblock Bayesian deep learning via subnetwork inference.
\newblock \emph{International Conference on Machine Learning (ICML)}, 2021.

\bibitem[De~Bortoli et~al.(2024)De~Bortoli, Hutchinson, Wirnsberger, and Doucet]{de2024target}
De~Bortoli, V., Hutchinson, M., Wirnsberger, P., and Doucet, A.
\newblock Target score matching.
\newblock \emph{arXiv preprint arXiv:2402.08667}, 2024.

\bibitem[Devroye et~al.(1996)Devroye, Gy{\"o}rfi, and Lugosi]{devroye1996probabilistic}
Devroye, L., Gy{\"o}rfi, L., and Lugosi, G.
\newblock \emph{A Probabilistic Theory of Pattern Recognition}.
\newblock Springer, 1996.

\bibitem[Doan et~al.(2025)Doan, Shamsi, Guo, Mohammadi, Alinejad-Rokny, Sejdinovic, Teney, Ranasinghe, and Abbasnejad]{doan2025bayesian}
Doan, B.~G., Shamsi, A., Guo, X.-Y., Mohammadi, A., Alinejad-Rokny, H., Sejdinovic, D., Teney, D., Ranasinghe, D.~C., and Abbasnejad, E.
\newblock Bayesian low-rank learning (bella): A practical approach to {Bayesian} neural networks.
\newblock \emph{Association for the Advancement of Artificial Intelligence (AAAI)}, 2025.

\bibitem[Dong et~al.(2022)Dong, Li, Dai, Zheng, Ma, Li, Xia, Xu, Wu, Liu, et~al.]{dong2022survey}
Dong, Q., Li, L., Dai, D., Zheng, C., Ma, J., Li, R., Xia, H., Xu, J., Wu, Z., Liu, T., et~al.
\newblock A survey on in-context learning.
\newblock \emph{arXiv preprint arXiv:2301.00234}, 2022.

\bibitem[Doob(1949)]{doob1949}
Doob, J.
\newblock Application of the theory of martingales.
\newblock \emph{Colloque International Centre Nat. Rech. Sci.}, pp.\  22--28, 1949.

\bibitem[Draxler et~al.(2018)Draxler, Veschgini, Salmhofer, and Hamprecht]{draxler2019essentially}
Draxler, F., Veschgini, K., Salmhofer, M., and Hamprecht, F.~A.
\newblock Essentially no barriers in neural network energy landscape.
\newblock \emph{International Conference on Machine Learning (ICML)}, 2018.

\bibitem[Elmoznino et~al.(2024)Elmoznino, Marty, Kasetty, Gagnon, Mittal, Fathi, Sridhar, and Lajoie]{elmoznino2024context}
Elmoznino, E., Marty, T., Kasetty, T., Gagnon, L., Mittal, S., Fathi, M., Sridhar, D., and Lajoie, G.
\newblock In-context learning and occam's razor.
\newblock \emph{arXiv preprint arXiv:2410.14086}, 2024.

\bibitem[Falck et~al.(2024)Falck, Wang, and Holmes]{falck2024incontext}
Falck, F., Wang, Z., and Holmes, C.
\newblock Is in-context learning in large language models {Bayesian}? a martingale perspective.
\newblock \emph{International Conference on Machine Learning (ICML)}, 2024.

\bibitem[Ferbach et~al.(2024)Ferbach, Goujaud, Gidel, and Dieuleveut]{ferbach2024proving}
Ferbach, D., Goujaud, B., Gidel, G., and Dieuleveut, A.
\newblock Proving linear mode connectivity of neural networks via optimal transport.
\newblock \emph{Artificial Intelligence and Statistics (AISTATS)}, 2024.

\bibitem[Fischer et~al.(2023)Fischer, Feurer, and Bischl]{fischer2023openmlctr23}
Fischer, S.~F., Feurer, M., and Bischl, B.
\newblock Open{ML}-{CTR}23 {\textendash} a curated tabular regression benchmarking suite.
\newblock In \emph{AutoML Conference 2023 (Workshop)}, 2023.
\newblock URL \url{https://openreview.net/forum?id=HebAOoMm94}.

\bibitem[Garg et~al.(2022)Garg, Tsipras, Liang, and Valiant]{garg2022can}
Garg, S., Tsipras, D., Liang, P.~S., and Valiant, G.
\newblock What can transformers learn in-context? a case study of simple function classes.
\newblock \emph{Neural Information Processing Systems (NeurIPS)}, 2022.

\bibitem[Garnelo et~al.(2018)Garnelo, Schwarz, Rosenbaum, Viola, Rezende, Eslami, and Teh]{garnelo2018neural}
Garnelo, M., Schwarz, J., Rosenbaum, D., Viola, F., Rezende, D.~J., Eslami, S., and Teh, Y.~W.
\newblock Neural processes.
\newblock \emph{arXiv preprint arXiv:1807.01622}, 2018.

\bibitem[Gilks \& Roberts(1996)Gilks and Roberts]{gilks1996strategies}
Gilks, W.~R. and Roberts, G.~O.
\newblock Strategies for improving {MCMC}.
\newblock \emph{Markov chain Monte Carlo in practice}, 6:\penalty0 89--114, 1996.

\bibitem[Hinton et~al.(1995)Hinton, Dayan, Frey, and Neal]{hinton1995wake}
Hinton, G.~E., Dayan, P., Frey, B.~J., and Neal, R.~M.
\newblock The ``wake-sleep'' algorithm for unsupervised neural networks.
\newblock \emph{Science}, 268 5214:\penalty0 1158--61, 1995.

\bibitem[Ho et~al.(2020)Ho, Jain, and Abbeel]{ho2020denoising}
Ho, J., Jain, A., and Abbeel, P.
\newblock Denoising diffusion probabilistic models.
\newblock \emph{Neural Information Processing Systems (NeurIPS)}, 2020.

\bibitem[Hoffman \& Gelman(2014)Hoffman and Gelman]{hoffman2014no}
Hoffman, M.~D. and Gelman, A.
\newblock The no-u-turn sampler: adaptively setting path lengths in hamiltonian monte carlo.
\newblock \emph{Journal of Machine Learning Research}, 15\penalty0 (1):\penalty0 1593--1623, 2014.

\bibitem[Kingma(2014)]{kingma2013auto}
Kingma, D.~P.
\newblock Auto-encoding variational bayes.
\newblock \emph{International Conference on Learning Representations (ICLR)}, 2014.

\bibitem[Kingma et~al.(2016)Kingma, Salimans, Jozefowicz, Chen, Sutskever, and Welling]{kingma2016improved}
Kingma, D.~P., Salimans, T., Jozefowicz, R., Chen, X., Sutskever, I., and Welling, M.
\newblock Improved variational inference with inverse autoregressive flow.
\newblock \emph{Neural Information Processing Systems (NIPS)}, 2016.

\bibitem[Kobyzev et~al.(2020)Kobyzev, Prince, and Brubaker]{kobyzev2020normalizing}
Kobyzev, I., Prince, S.~J., and Brubaker, M.~A.
\newblock Normalizing flows: An introduction and review of current methods.
\newblock \emph{IEEE transactions on pattern analysis and machine intelligence}, 43\penalty0 (11):\penalty0 3964--3979, 2020.

\bibitem[Lipman et~al.(2023)Lipman, Chen, Ben-Hamu, Nickel, and Le]{lipman2022flow}
Lipman, Y., Chen, R.~T., Ben-Hamu, H., Nickel, M., and Le, M.
\newblock Flow matching for generative modeling.
\newblock \emph{International Conference on Learning Representations (ICLR)}, 2023.

\bibitem[Miller(2018)]{miller2018detailed}
Miller, J.~W.
\newblock A detailed treatment of {Doob’s} theorem.
\newblock \emph{arXiv preprint arXiv:1801.03122}, 2018.

\bibitem[Miller(2021)]{miller2021asymptotic}
Miller, J.~W.
\newblock Asymptotic normality, concentration, and coverage of generalized posteriors.
\newblock \emph{Journal of Machine Learning Research}, 22\penalty0 (168):\penalty0 1--53, 2021.

\bibitem[Mittal et~al.(2023)Mittal, Bracher, Lajoie, Jaini, and Brubaker]{mittal2023exploring}
Mittal, S., Bracher, N.~L., Lajoie, G., Jaini, P., and Brubaker, M.~A.
\newblock Exploring exchangeable dataset amortization for bayesian posterior inference.
\newblock In \emph{ICML 2023 Workshop on Structured Probabilistic Inference $\{$$\backslash$\&$\}$ Generative Modeling}, 2023.

\bibitem[Mittal et~al.(2024)Mittal, Elmoznino, Gagnon, Bhardwaj, Sridhar, and Lajoie]{mittal2024does}
Mittal, S., Elmoznino, E., Gagnon, L., Bhardwaj, S., Sridhar, D., and Lajoie, G.
\newblock Does learning the right latent variables necessarily improve in-context learning?
\newblock \emph{arXiv preprint arXiv:2405.19162}, 2024.

\bibitem[Neal(1996)]{neal1996bayesian}
Neal, R.~M.
\newblock \emph{Bayesian learning for neural networks}.
\newblock Springer, 1996.

\bibitem[Neal(2011)]{neal2012mcmc}
Neal, R.~M.
\newblock {MCMC} using {Hamiltonian} dynamics.
\newblock \emph{Handbook of Markov Chain Monte Carlo}, 2\penalty0 (11):\penalty0 2, 2011.

\bibitem[Nichol \& Dhariwal(2021)Nichol and Dhariwal]{nichol2021improved}
Nichol, A.~Q. and Dhariwal, P.
\newblock Improved denoising diffusion probabilistic models.
\newblock \emph{International Conference on Machine Learning (ICML)}, 2021.

\bibitem[Papamakarios et~al.(2021)Papamakarios, Nalisnick, Rezende, Mohamed, and Lakshminarayanan]{papamakarios2021normalizing}
Papamakarios, G., Nalisnick, E., Rezende, D.~J., Mohamed, S., and Lakshminarayanan, B.
\newblock Normalizing flows for probabilistic modeling and inference.
\newblock \emph{Journal of Machine Learning Research}, 22\penalty0 (57):\penalty0 1--64, 2021.

\bibitem[Pearson(1936)]{pearson1936method}
Pearson, K.
\newblock Method of moments and method of maximum likelihood.
\newblock \emph{Biometrika}, 28\penalty0 (1/2):\penalty0 34--59, 1936.

\bibitem[Radev et~al.(2020)Radev, Mertens, Voss, Ardizzone, and K{\"o}the]{radev2020bayesflow}
Radev, S.~T., Mertens, U.~K., Voss, A., Ardizzone, L., and K{\"o}the, U.
\newblock Bayesflow: Learning complex stochastic models with invertible neural networks.
\newblock \emph{IEEE transactions on neural networks and learning systems}, 33\penalty0 (4):\penalty0 1452--1466, 2020.

\bibitem[Rainforth et~al.(2018)Rainforth, Kosiorek, Le, Maddison, Igl, Wood, and Teh]{rainforth2018tighter}
Rainforth, T., Kosiorek, A., Le, T.~A., Maddison, C., Igl, M., Wood, F., and Teh, Y.~W.
\newblock Tighter variational bounds are not necessarily better.
\newblock \emph{International Conference on Machine Learning (ICML)}, 2018.

\bibitem[Rezende \& Mohamed(2015)Rezende and Mohamed]{rezende2015variational}
Rezende, D. and Mohamed, S.
\newblock Variational inference with normalizing flows.
\newblock \emph{International Conference on Machine Learning (ICML)}, 2015.

\bibitem[Rezende et~al.(2014)Rezende, Mohamed, and Wierstra]{rezende2014stochastic}
Rezende, D.~J., Mohamed, S., and Wierstra, D.
\newblock Stochastic backpropagation and variational inference in deep latent {Gaussian} models.
\newblock \emph{International Conference on Machine Learning (ICML)}, 2014.

\bibitem[Ritter et~al.(2018)Ritter, Botev, and Barber]{ritter2018scalable}
Ritter, H., Botev, A., and Barber, D.
\newblock A scalable laplace approximation for neural networks.
\newblock \emph{International Conference on Representation Learning (ICLR)}, 2018.

\bibitem[S{\"a}rkk{\"a} \& Solin(2019)S{\"a}rkk{\"a} and Solin]{sarkka2019applied}
S{\"a}rkk{\"a}, S. and Solin, A.
\newblock \emph{Applied stochastic differential equations}, volume~10.
\newblock Cambridge University Press, 2019.

\bibitem[Sendera et~al.(2024)Sendera, Kim, Mittal, Lemos, Scimeca, Rector-Brooks, Adam, Bengio, and Malkin]{sendera2024improved}
Sendera, M., Kim, M., Mittal, S., Lemos, P., Scimeca, L., Rector-Brooks, J., Adam, A., Bengio, Y., and Malkin, N.
\newblock Improved off-policy training of diffusion samplers.
\newblock \emph{Neural Information Processing Systems (NeurIPS)}, 2024.

\bibitem[Sharma et~al.(2023)Sharma, Farquhar, Nalisnick, and Rainforth]{sharma2023bayesian}
Sharma, M., Farquhar, S., Nalisnick, E., and Rainforth, T.
\newblock Do {Bayesian} neural networks need to be fully stochastic?
\newblock \emph{Artificial Intelligence and Statistics (AISTATS)}, 2023.

\bibitem[Song et~al.(2021{\natexlab{a}})Song, Meng, and Ermon]{song2020denoising}
Song, J., Meng, C., and Ermon, S.
\newblock Denoising diffusion implicit models.
\newblock \emph{International Conference on Learning Representations (ICLR)}, 2021{\natexlab{a}}.

\bibitem[Song \& Ermon(2020)Song and Ermon]{song2020improved}
Song, Y. and Ermon, S.
\newblock Improved techniques for training score-based generative models.
\newblock \emph{Neural Information Processing Systems (NeurIPS)}, 2020.

\bibitem[Song et~al.(2021{\natexlab{b}})Song, Sohl-Dickstein, Kingma, Kumar, Ermon, and Poole]{song2020score}
Song, Y., Sohl-Dickstein, J., Kingma, D.~P., Kumar, A., Ermon, S., and Poole, B.
\newblock Score-based generative modeling through stochastic differential equations.
\newblock \emph{International Conference on Learning Representations (ICLR)}, 2021{\natexlab{b}}.

\bibitem[Teicher(1963)]{teicher1963identifiability}
Teicher, H.
\newblock Identifiability of finite mixtures.
\newblock \emph{The Annals of Mathematical statistics}, pp.\  1265--1269, 1963.

\bibitem[Tong et~al.(2024)Tong, Fatras, Malkin, Huguet, Zhang, Rector-Brooks, Wolf, and Bengio]{tong2023improving}
Tong, A., Fatras, K., Malkin, N., Huguet, G., Zhang, Y., Rector-Brooks, J., Wolf, G., and Bengio, Y.
\newblock Improving and generalizing flow-based generative models with minibatch optimal transport.
\newblock \emph{Transactions on Machine Learning Research}, 2024.

\bibitem[Vargas et~al.(2023)Vargas, Grathwohl, and Doucet]{vargas2023denoising}
Vargas, F., Grathwohl, W., and Doucet, A.
\newblock Denoising diffusion samplers.
\newblock \emph{International Conference on Learning Representations (ICLR)}, 2023.

\bibitem[Von~Oswald et~al.(2023)Von~Oswald, Niklasson, Randazzo, Sacramento, Mordvintsev, Zhmoginov, and Vladymyrov]{von2023transformers}
Von~Oswald, J., Niklasson, E., Randazzo, E., Sacramento, J., Mordvintsev, A., Zhmoginov, A., and Vladymyrov, M.
\newblock Transformers learn in-context by gradient descent.
\newblock \emph{International Conference on Machine Learning (ICML)}, 2023.

\bibitem[Welling \& Teh(2011)Welling and Teh]{welling2011bayesian}
Welling, M. and Teh, Y.~W.
\newblock Bayesian learning via stochastic gradient {Langevin} dynamics.
\newblock \emph{International Conference on Machine Learning (ICML)}, 2011.

\bibitem[Wenzel et~al.(2020)Wenzel, Roth, Veeling, Swiatkowski, Tran, Mandt, Snoek, Salimans, Jenatton, and Nowozin]{wenzel2020good}
Wenzel, F., Roth, K., Veeling, B.~S., Swiatkowski, J., Tran, L., Mandt, S., Snoek, J., Salimans, T., Jenatton, R., and Nowozin, S.
\newblock How good is the {Bayes} posterior in deep neural networks really?
\newblock \emph{International Conference on Machine Learning (ICML)}, 2020.

\bibitem[Xie et~al.(2022)Xie, Raghunathan, Liang, and Ma]{xie2021explanation}
Xie, S.~M., Raghunathan, A., Liang, P., and Ma, T.
\newblock An explanation of in-context learning as implicit {Bayesian} inference.
\newblock \emph{International Conference on Learning Representations (ICLR)}, 2022.

\bibitem[Yakowitz \& Spragins(1968)Yakowitz and Spragins]{yakowitz1968identifiability}
Yakowitz, S.~J. and Spragins, J.~D.
\newblock On the identifiability of finite mixtures.
\newblock \emph{The Annals of Mathematical Statistics}, 39\penalty0 (1):\penalty0 209--214, 1968.

\bibitem[Zhang \& Chen(2022)Zhang and Chen]{zhang2021path}
Zhang, Q. and Chen, Y.
\newblock Path integral sampler: a stochastic control approach for sampling.
\newblock \emph{International Conference on Learning Representations (ICLR)}, 2022.

\end{thebibliography}
\bibliographystyle{bibstyle}

\appendix
\onecolumn
\section{Related Work}

\paragraph{Normalizing Flows}. Since Gaussian distributions are unimodal, they cannot approximate more complex multi-modal distributions well. To alleviate this problem, multiple works start with a simple distribution and then apply learnable invertible transformations to construct more complicated densities~\citep{rezende2015variational,kobyzev2020normalizing,papamakarios2021normalizing,kingma2016improved}. These transformations are designed in a manner that the jacobian determinant is easily computable to allow for ease in computing entropy and training via back-propagation.
\begin{align}
    q^j_\varphi(\cdot) &= g_j \circ \ldots \circ g_1 \circ \gN(\cdot; \bf{0}, \bf{I}) \\
    q_\varphi^j(\theta^j) &= q^{j-1}_\varphi(g_j^{-1}(\theta^j)) |J_{g_j}(g_j^{-1}(\theta^j))|^{-1}
\end{align}
for $j = 1, \ldots, n$ and $q_\varphi^0$ represents the standard normal distribution. Further, $J_{g_j}$ represents the jacobian of the invertible function $g_j$ and $|\cdot|$ represents the determinant operator.

\paragraph{Score-Based Generative Modeling}. Recent advances in generative models have stemmed from diffusion models~\citep{song2020score,song2020improved,song2020denoising,ho2020denoising,nichol2021improved} that consider a forward noising process via a stochastic differential equation as 
\begin{align}
    d\theta_t = f(\theta_t, t) \;dt + g(t)\;d\text{w}_t
\end{align}
with the corresponding reverse process as~\citep{anderson1982reverse}
\begin{align}
    d\theta_t = f(\theta_t, t) - g(t)^2 \nabla_\theta \log p_t(\theta)|_{\theta_t} + g(t) d\bar{\text{w}}_t
\end{align}
Note that this requires estimating the score function at all time-steps $t$ to integrate the SDE and obtain samples. Prior work has shown that the denoising objective provides a viable method for obtaining an estimate of the score function provided access to data, which is trained as
\begin{align}
    \arg\min_\varphi \mathbb{E}_{t, \theta_0, \theta_t} \left[
    \norm{s_\varphi(\theta_t, t) - \nabla_{\theta_t} \log p(\theta_t | \theta_0)}^2\right]
\end{align}
Note that if $f$ is a linear function, one can sample $\theta_t$ given $\theta_0$ and $t$ directly in a simulation-free manner~\citep{sarkka2019applied}, which allows for scalable training of diffusion models through the above equation.

\textbf{Flow-Matching}. Contrary to diffusion models, flow-matching~\citep{lipman2022flow,tong2023improving} models data through an ordinary differential equation instead of a stochastic differential equation. It first constructs an interpolation~\citep{albergo2023stochastic,albergo2023dependent}, possibly noisy, between two random variables $\theta_0$ and $\theta_1$ as
\begin{align}
    \theta_t = \alpha_t \theta_0 + \beta_t \theta_1 + \gamma_t \vz
\end{align}
where $\alpha_0 = \beta_1 = 1$, $\alpha_1 = \beta_0 = 0$ and $\gamma_0 = \gamma_1 = 0$, and $\vz$ follows a normal distribution. Samples from the target density can then be obtained by sampling a $\theta_1$ and then solving the following ODE dynamics
\begin{align}
     d\theta_t = v_\varphi(\theta_t, t)\,dt
\end{align}
where the maginal drift is trained as
\begin{align}
    \arg\min_\varphi \mathbb{E}_{\theta_0, \theta_1, t, \vz, \theta_t}\left[\norm{v_\varphi(\theta_t, t) - \partial_t \theta_t}^2\right]
\end{align}

\textbf{Denoising Energy Matching}. It is important to note that diffusion models are trained via data, while multiple applications require training a model to sample proportional to an unnormalized distribution in the absence of any data. Denoising Energy Matching (DEM)~\citep{akhound2024iterated} provides an importance sampling based estimate to train a similar diffusion model in the absence of data, by considering the target score matching estimator~\citep{de2024target} and combining it with importance sampling with the transition kernel $p(\theta_t | \theta_0)$ as the proposal for $\theta_0$.

\section{Probabilistic Models}
\label{apdx:probabilistic_models}
We discuss various probabilistic models used in our experiments as well as the form for the likelihood and the prior. This closely follows the setup in \citet{mittal2023exploring}.

\textbf{Mean of Gaussian (GM):} We consider estimating the mean $\mmu$ of a Gaussian distribution given some observed data. In this case, prior and likelihood defining the probabilistic model $p(\vx, \mtheta)$ (with $\mtheta$ being the mean $\mmu$) are given by:
\begin{align}
    p(\mmu) &= \gN\left(\mmu | \mathbf{0}, \mathbf{I}\right)\\
    p(\vx | \mmu) &= \gN\left(\vx | \mmu, \mSigma\right) 
\end{align}
and $\mSigma$ is known beforehand and defined as a unit variance matrix. 

\textbf{Linear Regression (LR):} We estimate the weight vector for Bayesian linear regression, where the underlying model $p(\gD, \mtheta)$ is given by:
\begin{align}
    p(\vw) &= \gN(\vw | \mathbf{0}, \mathbf{I})\\
    p(b) &= \gN(b | 0, 1)\\
    p(y | \vx, \vw, b) &= \gN\left(y | \vw^T\vx + b, \sigma^2\right) \, ,
\end{align}
and with $\sigma^2 = 0.25$ known beforehand. Inputs $\vx$ are generated from $p(\vx) = \gN(\mathbf{0}, I)$.

\textbf{Linear Classification (LC):}
The underlying probabilistic model is:
\begin{align}
    p(\mW) &= \gN\left(\mW | \mathbf{0}, \mathbf{I}\right)\\
    p(y | \vx, \mW) &= \mathrm{Categorical}\left(y  \;\vline\; \frac{1}{\tau}\;\mW\vx\right)\, ,
\end{align}
where $\tau$ is the known temperature term which is kept as $0.1$ to ensure peaky distributions, and $\vx$ is being generated from $p(\vx) = \gN(\mathbf{0}, I)$.

\textbf{Nonlinear Regression (NLR):}
We consider the model as a Bayesian Neural Network (BNN) for regression with fixed hyper-parameters like the number of layers, dimensionality of the hidden layer, etc. Let the BNN denote the function $f_\mtheta$ where $\mtheta$ are the network parameters. Then, for regression, we specify the probabilistic model using:
\begin{align}
    p(\mtheta) &= \gN\left(\mtheta | \mathbf{0}, \mathbf{I}\right)\\
    p(y | \vx, \mtheta) &= \gN\left(y | f_\mtheta(\vx), \sigma^2\right) \, ,
\end{align}
where $\sigma^2 = 0.25$ is a known quantity and $\vx$ being generated from $p(\vx) = \gN(\mathbf{0}, I)$.
 
\textbf{Nonlinear Classification (NLC):}
Like in Nonlinear Regression, we consider BNNs with fixed hyper-parameters for classification problems with the same estimation task. In this formulation, we consider the probabilistic model as:
\begin{align}
    p(\mtheta) &= \gN\left(\mtheta | \mathbf{0}, \mathbf{I}\right)\\
    p(y | \vx, \mtheta) &= \mathrm{Categorical}\left(y \;\vline\; \frac{1}{\tau}\;f_\mtheta(\vx)\right)
\end{align}
where $\tau$ is the known temperature term which is kept as $0.1$ to ensure peaky distributions, and $\vx$ is being generated from $p(\vx) = \gN(\mathbf{0}, I)$.

\textbf{Gaussian Mixture Model (GMM):}
We look at a well-known probabilistic model for unsupervised learning, Gaussian Mixture Model (GMM), primarily used to cluster data. Consider a $K$-cluster GMM with:
\begin{align}
    p(\mmu_k) &= \gN\left(\mmu_k | \mathbf{0}, \mathbf{I}\right)\\
    p(\vx | \mmu_{1:K}) &= \sum_{k=1}^K \pi_k \gN\left(\vx | \mmu_k, \mSigma_k\right) \, .
\end{align}
 We assume $\mSigma_k$ and $\pi_k$ to be known and set $\mSigma_k$ to be an identity matrix and the mixing coefficients to be equal, $\pi_k = 1/K$, for all clusters $k$ in our experiments. 

\section{Metrics}
\label{apdx:metrics}
\textbf{Regression}. For regression problems, we consider the ensembled loss metric as the following
\begin{align}
    \mathbb{E}_{\vx, \vy, \gD} \left[\norm{\vy - \mathbb{E}_{q_\varphi(\theta | \gD)}\left[\hat{\vy} \Big\rvert \vx, \theta\right]}^2\right]
\end{align}
while the single-sample metric is defined as
\begin{align}
    \mathbb{E}_{\vx, \vy, \gD} \mathbb{E}_{q_\varphi(\theta | \gD)} \left[\norm{\vy - \hat{\vy}}^2 \Big\rvert \vx, \theta\right]
\end{align}
where $\hat{\vy}$ denotes the mode of the distribution $p(\vy | \vx, \theta)$. We rely on similar metrics for the estimation of the mean of a Gaussian distribution, with the only difference being the absence of $\vx$, and $\hat{y} = \theta$, as it is an unsupervised learning problem.

\textbf{Classification}. For classification problems, we consider the ensembled accuracy metric which is obtained as the following
\begin{align}
    100 \times \mathbb{E}_{\vx, \vy, \gD} \left[\mathbbm{1}_\vy\left(\text{Mode}\left(\hat{\vy}_1, \ldots, \hat{\vy}_s\right)\right)\right]
\end{align}
where $\mathbbm{1}_\vy(\cdot)$ is an indicator function which is $1$ if the argument is the same as $\vy$ and $0$ otherwise. Mode represents the mode of its arguments, where each $\hat{\vy}_i$ is the mode of  $p(\vy | \vx, \theta_i)$ with $\theta_i \sim q_\varphi(\theta | \gD)$. Similarly, the single sample metric is defined as
\begin{align}
    100 \times \mathbb{E}_{\vx, \vy, \gD} \mathbb{E}_{q_\varphi(\theta | \gD)} \left[\mathbbm{1}_\vy\left(\hat{\vy}\right)\Big\rvert \vx, \theta\right]
\end{align}
Note that the multiplication by $100$ is just to scale the accuracy to $0-100$.

\textbf{Gaussian Mixture Model}. For the Gaussian Mixture Model, there is no clear notion of ensembling due to the identifiability problem in clustering, \ie, averaging over two clusters could lead to the average not corresponding to any meaningful cluster. Thus, we only consider single sample metric for this case, in particular
\begin{align}
    \mathbb{E}_{\vy, \gD} \mathbb{E}_{q_\varphi(\theta_1, \ldots \theta_c | \gD)} \left[\left(\vy - \argmin_{\psi \in \theta_1, \ldots \theta_c} \,\left(\vy - \psi\right)^2\right)\right]
\end{align}
where $\theta_1, \ldots \theta_c$ can be subsumed into a single larger vector $\theta$ for the purposes of modeling a $q_\varphi$.

Note that in case of point estimates, for all the metrics, $q_\varphi$ can be considered as a dirac measure and both ensembled and single-sample metrics represent the same quantity.

\section{Implementation Details}
\label{apdx:implementation}
In this section, we outline the implementation details behind each of the estimators. We consider the transformer architecture in all cases to model the conditioning on the set of observations $\gD$. We remove the positional embeddings so that the inferred parameters, distribution or point, are permutation invariant to $\gD$. We use [CLS] as an additional token embedded to the sequence and the prediction corresponding to it is used to infer the parameters.

For the architecture details, we use $4$ encoder layers with a $256$ dimensional attention block and $1024$ feed-forward dimensions, and $4$ heads in each attention block for our Transformer models.

We use a diagonal Gaussian assumption for modeling densities using the Gaussian distribution.
For discrete normalizing flows, we follow the setup in \cite{radev2020bayesflow} and use $6$ coupling blocks, each with a $1$ hidden-layered non-linear feed-forward subnetwork with ReLU non-linearity and $128$ hidden dimensions.

For score-based diffusion models, we use the variance exploding SDE with no drift and the diffusion coefficient $g_t$ set to be $\sqrt{2t\beta^2}$ and train the estimator using denoising score matching with the loss being equally weighted for all times $t$. We use the same schedule for pDEM as well, and use $100$ samples in the importance-sample estimate of the score.

Finally, we use linear interpolation scheme for flow-matching that interpolates between the parameters $\theta$ and unstructured noise $z$ in a linear manner.

For inference in continuous time models, we use $100$ steps to perform both the SDE and ODE integration.

For training the in-context estimators, we sample the number of observations $|\gD|$ randomly in the interval $[64, 128]$. For the variable-dimensional experiments, we randomly sample the problem dimensionality in the range $[1, 100]$. To evaluate the models, we sample $100$ different datasets maintaining the test set to be the same across different seeds.

All the in-context estimators are trained until convergence, in particular,

\textbf{GM}: $50k$ for fixed-dimensional and $100k$ for variable-dimensional.

\textbf{GMM}: $250k$ for fixed-dimensional and $500k$ for variable-dimensional.

\textbf{LR}: $150k$ for fixed-dimensional and $250k$ for variable-dimensional.

\textbf{NLR}: $250k$ for fixed-dimensional and $500k$ for variable-dimensional.

\textbf{LC}: $150k$ for fixed-dimensional and $250k$ for variable-dimensional.

\textbf{NLC}: $250k$ for fixed-dimensional and $500k$ for variable-dimensional.

In addition, all experiments on synthetic misspecification are trained for $250k$ iterations.

\begin{table*}[t]
    \centering
    \small
    % \scriptsize
    \def\arraystretch{1.05}
    \setlength{\tabcolsep}{7pt}
    \begin{tabular}{l r c c c c c c c}
        \toprule
          & & \multicolumn{4}{c}{\textit{$L_2$ Loss} ($\downarrow$)} & \multicolumn{2}{c}{\textit{Accuracy} ($\uparrow$)}\\
         \cmidrule(lr){3-6}\cmidrule(lr){7-8}
        & \textbf{Objective} & \multicolumn{1}{c}{\textbf{GM}} & \multicolumn{1}{c}{\textbf{GMM}} & \multicolumn{1}{c}{\textbf{LR}} & \multicolumn{1}{c}{\textbf{NLR}} & \multicolumn{1}{c}{\textbf{LC}} & \multicolumn{1}{c}{\textbf{NLC}} \\
        \cmidrule(lr){3-8}
        & & $100$D & $5$D $2$cl & $100$D & $25$D & $100$D $2$cl & $25$D $2$cl \\
\midrule
\multirow{4}{*}{Baseline} & Random & $301.06$\sstd{$0.35$} & $5.00$\sstd{$0.04$} & $202.6$\sstd{$0.3$} & $831.6$\sstd{$8.7$} & $50.0$\sstd{$0.0$} & $50.0$\sstd{$0.3$} \\
& Optimization & $101.24$\sstd{$0.00$} & $0.42$\sstd{$0.00$} & $25.1$\sstd{$0.0$} & $104.0$\sstd{$0.1$} & $70.3$\sstd{$0.0$} & $77.9$\sstd{$0.0$} \\
& Langevin & $102.35$\sstd{$0.03$} & $0.45$\sstd{$0.01$} & $23.3$\sstd{$0.7$} & $132.4$\sstd{$1.0$} & $65.1$\sstd{$0.4$} & $73.2$\sstd{$0.3$} \\
& HMC & $102.41$\sstd{$0.03$} & $0.48$\sstd{$0.01$} & $18.7$\sstd{$0.2$} & $98.1$\sstd{$0.7$} & $62.1$\sstd{$0.2$} & $70.4$\sstd{$0.1$} \\
\midrule
\multirow{3}{*}{Gaussian} & Fwd-KL &$102.78$\sstd{$0.00$} & $2.50$\sstd{$0.03$} & $45.9$\sstd{$1.3$} & $680.9$\sstd{$5.8$} & $63.0$\sstd{$0.1$} & $57.1$\sstd{$0.4$} \\

& Rev-KL &$102.54$\sstd{$0.03$} & \highlight{$0.49$\sstd{$0.02$}} & $28.7$\sstd{$0.3$} & $102.3$\sstd{$1.8$} & $68.2$\sstd{$0.0$} & $75.2$\sstd{$0.1$} \\

& Sym-KL &$102.63$\sstd{$0.03$} & $0.66$\sstd{$0.01$} & $31.4$\sstd{$0.2$} & $105.6$\sstd{$0.7$} & $66.8$\sstd{$0.1$} & $71.3$\sstd{$0.1$} \\
\midrule

\multirow{3}{*}{Norm. Flows} & Fwd-KL &$102.77$\sstd{$0.02$} & $0.62$\sstd{$0.07$} & $43.3$\sstd{$2.7$} & $539.3$\sstd{$4.3$} & $64.3$\sstd{$0.1$} & $58.3$\sstd{$0.1$} \\

& Rev-KL &$102.53$\sstd{$0.05$} & \highlight{$0.47$\sstd{$0.01$}} & \highlight{$29.4$\sstd{$1.6$}} & $102.6$\sstd{$0.9$} & $68.7$\sstd{$0.1$} & $75.0$\sstd{$0.5$} \\

& Sym-KL &$102.61$\sstd{$0.02$} & \highlight{$0.48$\sstd{$0.02$}} & $30.5$\sstd{$0.6$} & $104.8$\sstd{$0.4$} & $68.5$\sstd{$0.0$} & $74.3$\sstd{$0.9$} \\
\midrule

\multirow{3}{*}{Diffusion} & Score-Based &$103.00$\sstd{$0.04$} & $0.51$\sstd{$0.00$} & $39.3$\sstd{$0.4$} & $991.8$\sstd{$6.9$} & $63.1$\sstd{$0.5$} & $55.7$\sstd{$0.0$} \\

& Flow-Matching &$102.69$\sstd{$0.08$} & $0.61$\sstd{$0.02$} & $45.3$\sstd{$0.3$} & $659.9$\sstd{$3.2$} & $64.6$\sstd{$0.1$} & $57.9$\sstd{$0.2$} \\

& pDEM &$115.36$\sstd{$0.98$} & $0.61$\sstd{$0.02$} & $61.0$\sstd{$1.0$} & $307.6$\sstd{$4.4$} & $70.6$\sstd{$0.2$} & $67.0$\sstd{$0.2$} \\
\midrule

\multirow{2}{*}{Point} & MLE &$101.30$\sstd{$0.00$} & $0.49$\sstd{$0.01$} & \highlight{$28.1$\sstd{$0.7$}} & \highlight{$99.0$\sstd{$2.9$}} & $73.0$\sstd{$0.2$} & $76.5$\sstd{$0.4$} \\

& MAP & \highlight{$101.28$\sstd{$0.00$}} & $0.48$\sstd{$0.00$} & \highlight{$28.1$\sstd{$0.6$}} & \highlight{$96.9$\sstd{$1.5$}} & \highlight{$73.4$\sstd{$0.1$}} & \highlight{$78.3$\sstd{$0.2$}} \\
\bottomrule
    \end{tabular}
    \caption{\textbf{Fixed-Dimensional}. Evaluation of different in-context estimators under the single-sample metric for the suite of probabilistic models.}
    \vspace{-5mm}
    \label{tab:fixed_dim}
\end{table*}
\begin{table*}[t]
    \centering
    \small
    % \scriptsize
    \def\arraystretch{1.05}
    \setlength{\tabcolsep}{7pt}
    \begin{tabular}{l r c c c c c c}
        \toprule
          & & \multicolumn{4}{c}{\textit{$L_2$ Loss} ($\downarrow$)} & \multicolumn{2}{c}{\textit{Accuracy} ($\uparrow$)}\\
         \cmidrule(lr){3-6}\cmidrule(lr){7-8}
        & \textbf{Objective} & \multicolumn{1}{c}{\textbf{GM}} & \multicolumn{1}{c}{\textbf{GMM}} & \multicolumn{1}{c}{\textbf{LR}} & \multicolumn{1}{c}{\textbf{NLR}} & \multicolumn{1}{c}{\textbf{LC}} & \multicolumn{1}{c}{\textbf{NLC}} \\
        \cmidrule(lr){3-8}
        & & $100$D & $5$D $2$cl & $100$D & $50$D & $100$D $2$cl & $50$D $2$cl \\
\midrule
\multirow{4}{*}{Baseline} & Random & $298.46$\sstd{$0.44$} & $4.66$\sstd{$0.03$} & $200.4$\sstd{$0.7$} & $1696.8$\sstd{$11.9$} & $50.0$\sstd{$0.1$} & $49.9$\sstd{$0.3$} \\
& Optimization & $100.88$\sstd{$0.00$} & $0.43$\sstd{$0.00$} & $20.1$\sstd{$0.0$} & $309.2$\sstd{$0.2$} & $71.2$\sstd{$0.0$} & $76.1$\sstd{$0.1$} \\
& Langevin & $101.93$\sstd{$0.03$} & $0.44$\sstd{$0.00$} & $21.4$\sstd{$0.8$} & \textsc{N/A} & $65.4$\sstd{$0.4$} & $69.9$\sstd{$0.3$} \\
& HMC & $102.02$\sstd{$0.02$} & $0.46$\sstd{$0.01$} & $17.7$\sstd{$0.1$} & $303.3$\sstd{$2.4$} & $62.7$\sstd{$0.2$} & $68.2$\sstd{$0.4$} \\
\midrule

\multirow{3}{*}{Gaussian} & Fwd-KL &$108.93$\sstd{$0.10$} & $2.39$\sstd{$0.02$} & $63.8$\sstd{$2.7$} & $1320.0$\sstd{$11.1$} & $62.5$\sstd{$0.2$} & $58.9$\sstd{$0.3$} \\

& Rev-KL &$104.75$\sstd{$0.10$} & \highlight{$0.47$\sstd{$0.01$}} & $32.5$\sstd{$0.6$} & $279.6$\sstd{$2.6$} & $67.8$\sstd{$0.2$} & $73.6$\sstd{$0.3$} \\

& Sym-KL &$104.96$\sstd{$0.10$} & $0.63$\sstd{$0.02$} & $34.4$\sstd{$1.1$} & $284.3$\sstd{$0.7$} & $66.6$\sstd{$0.1$} & $70.9$\sstd{$0.2$} \\
\midrule

\multirow{3}{*}{Norm. Flows} & Fwd-KL &$108.56$\sstd{$0.16$} & $0.55$\sstd{$0.07$} & $61.4$\sstd{$2.3$} & $1078.5$\sstd{$7.8$} & $63.6$\sstd{$0.1$} & $60.3$\sstd{$0.1$} \\

& Rev-KL &$104.89$\sstd{$0.09$} & \highlight{$0.46$\sstd{$0.01$}} & $33.2$\sstd{$0.6$} & \highlight{$275.0$\sstd{$2.2$}} & $68.1$\sstd{$0.2$} & $72.6$\sstd{$0.2$} \\

& Sym-KL &$105.12$\sstd{$0.07$} & \highlight{$0.48$\sstd{$0.02$}} & $45.3$\sstd{$5.6$} & \highlight{$276.2$\sstd{$4.8$}} & $67.7$\sstd{$0.2$} & $64.4$\sstd{$0.5$} \\
\midrule

\multirow{3}{*}{Diffusion} & Score-Based &$106.79$\sstd{$0.09$} & $0.50$\sstd{$0.00$} & $47.0$\sstd{$1.2$} & $2668.1$\sstd{$19.4$} & $62.7$\sstd{$0.5$} & $57.5$\sstd{$0.3$} \\

& Flow-Matching &$106.83$\sstd{$0.22$} & $0.64$\sstd{$0.01$} & $52.0$\sstd{$1.3$} & $1206.2$\sstd{$11.2$} & $63.8$\sstd{$0.4$} & $60.3$\sstd{$0.1$} \\

& pDEM &$119.78$\sstd{$0.65$} & $0.66$\sstd{$0.13$} & $108.2$\sstd{$10.7$} & $654.8$\sstd{$11.5$} & $69.6$\sstd{$0.2$} & $65.8$\sstd{$0.2$} \\
\midrule

\multirow{2}{*}{Point} & MLE &$103.07$\sstd{$0.20$} & \highlight{$0.47$\sstd{$0.02$}} & \highlight{$31.4$\sstd{$0.4$}} & $289.1$\sstd{$3.2$} & $70.9$\sstd{$0.2$} & $75.5$\sstd{$0.2$} \\

& MAP & \highlight{$102.60$\sstd{$0.07$}} & \highlight{$0.47$\sstd{$0.02$}} & \highlight{$31.2$\sstd{$0.5$}} & $285.7$\sstd{$2.0$} & \highlight{$72.3$\sstd{$0.2$}} & \highlight{$76.3$\sstd{$0.2$}} \\
\bottomrule
    \end{tabular}
    \caption{\textbf{Variable-Dimensional}. Evaluation of different in-context estimators under the single-sample metric for the suite of probabilistic models.}
    \vspace{-5mm}
    \label{tab:variable_dim}
\end{table*}
\begin{table}[t]
    \centering
    \small
    \def\arraystretch{1.05}
    \setlength{\tabcolsep}{5pt}
    \begin{tabular}{ll|ccc}
    \toprule
     \multirow{2}{*}{$\chi_{test}\;(\rightarrow)$} & \multirow{2}{*}{\textit{Data}} & \multirow{2}{*}{Linear} & MLP & GP \\
     & & & Nonlinear & Nonlinear \\
    \cmidrule{2-5}
    $\chi_{train}\;(\rightarrow)$&  \textit{Model} & NLR & LR & NLR \\
    \midrule    
\multirow{4}{*}{Gaussian} & Fwd-KL & $15.298$\sstd{$0.308$} & $2.390$\sstd{$0.266$} & $14.672$\sstd{$0.441$} \\
& Sym-KL & $1.326$\sstd{$0.051$} & $1.799$\sstd{$0.115$} & $1.082$\sstd{$0.045$} \\
& Rev-KL  &$0.384$\sstd{$0.005$} & $2.386$\sstd{$0.838$} & $0.157$\sstd{$0.008$} \\
& \quad \textit{+ switched data} &$0.366$\sstd{$0.005$} & \highlight{$1.226$\sstd{$0.002$}} & $0.066$\sstd{$0.003$} \\
\midrule

\multirow{4}{*}{Norm. Flows} & Fwd-KL  &$7.922$\sstd{$0.349$} & $1.722$\sstd{$0.130$} & $8.462$\sstd{$0.479$} \\
& Sym-KL &$0.351$\sstd{$0.006$} & $1.434$\sstd{$0.065$} & $0.119$\sstd{$0.006$} \\
& Rev-KL &$0.352$\sstd{$0.007$} & $1.513$\sstd{$0.122$} & $0.123$\sstd{$0.006$} \\
& \quad \textit{+ switched data} &$0.343$\sstd{$0.006$} & \highlight{$1.226$\sstd{$0.004$}} & $0.057$\sstd{$0.002$} \\
\midrule

\multirow{4}{*}{Diffusion} & Score-Based &$2.647$\sstd{$0.094$} & $1.562$\sstd{$0.107$} & $2.430$\sstd{$0.044$} \\
& Flow-Matching &$6.755$\sstd{$0.812$} & $2.135$\sstd{$0.125$} & $6.199$\sstd{$0.776$} \\
& pDEM &$7.003$\sstd{$0.266$} & $1.734$\sstd{$0.169$} & $6.795$\sstd{$0.172$} \\
& \quad \textit{+ switched data} &$6.683$\sstd{$0.162$} & $1.691$\sstd{$0.701$} & $11.753$\sstd{$2.274$} \\
\midrule

\multirow{4}{*}{Point} & MLE &$0.399$\sstd{$0.019$} & $1.525$\sstd{$0.046$} & $0.027$\sstd{$0.016$} \\
& \quad \textit{+ switched data} &$0.382$\sstd{$0.007$} & \highlight{$1.227$\sstd{$0.003$}} & \highlight{$0.002$\sstd{$0.000$}} \\
& MAP &$0.267$\sstd{$0.001$} & $1.541$\sstd{$0.081$} & $0.025$\sstd{$0.000$} \\
& \quad \textit{+ switched data} & \highlight{$0.263$\sstd{$0.000$}} & \highlight{$1.225$\sstd{$0.004$}} & $0.014$\sstd{$0.000$} \\
\bottomrule
    \end{tabular}
    % \vspace{-2mm}
    \caption{\textbf{Model Misspecification}. Evaluation of the in-context estimators through single-sample $L_2$ metric in OoD generalization when the assumed model is misspecified, \ie the data of interest comes from $\chi_{test}$ but our modeling assumption corresponds to $\chi_{train}$.}
    % \vspace{-7mm}
    \label{tab:misspecification}
\end{table}
\begin{table}[t]
    \centering
    \small
    \def\arraystretch{1.05}
    \setlength{\tabcolsep}{5pt}
    \begin{tabular}{lc cccc}
        \toprule
         &  & \multicolumn{2}{c}{\textit{$L_2$ Loss} ($\downarrow)$} & \multicolumn{2}{c}{\textit{Accuracy} ($\uparrow$)}\\
        \cmidrule(lr){3-4}\cmidrule(lr){5-6}
        & \textbf{Model} & LR & NLR & LC & NLC \\
        \midrule
Random & - & $23.20$\sstd{$0.45$} & $212.25$\sstd{$9.54$} & $50.06$\sstd{$0.22$} & $50.93$\sstd{$0.86$} \\
\midrule
Fwd-KL & \multirow{3}{*}{\rotatebox[origin=c]{90}{Gaussian}} &$9.69$\sstd{$0.91$} & $116.97$\sstd{$8.97$} & $64.26$\sstd{$4.89$} & $59.46$\sstd{$1.45$} \\
Rev-KL & &$7.70$\sstd{$0.68$} & $8.02$\sstd{$1.06$} & $74.45$\sstd{$4.84$} & $72.33$\sstd{$5.04$} \\
Sym-KL & &$7.48$\sstd{$0.52$} & $26.33$\sstd{$18.18$} & $73.08$\sstd{$2.41$} & $72.05$\sstd{$2.32$} \\
\midrule
Fwd-KL & \multirow{3}{*}{\rotatebox[origin=c]{90}{Flow}} &$9.60$\sstd{$0.38$} & $81.40$\sstd{$8.39$} & $68.59$\sstd{$2.86$} & $60.64$\sstd{$1.74$} \\
Rev-KL & &$7.59$\sstd{$0.49$} & \highlight{$8.71$\sstd{$1.70$}} & $73.19$\sstd{$5.40$} & \highlight{$76.72$\sstd{$2.65$}} \\
Sym-KL & &$7.53$\sstd{$0.43$} & \highlight{$10.07$\sstd{$4.70$}} & $76.45$\sstd{$2.79$} & $75.61$\sstd{$2.89$} \\
\midrule
Score-Based & \multirow{3}{*}{\rotatebox[origin=c]{90}{Diffusion}} & \highlight{$10.12$\sstd{$4.59$}} & $512.86$\sstd{$20.31$} & $60.93$\sstd{$8.68$} & $57.55$\sstd{$0.91$} \\
Flow-Matching & &$12.24$\sstd{$5.36$} & $158.85$\sstd{$5.77$} & $71.79$\sstd{$3.05$} & $61.74$\sstd{$0.88$} \\
pDEM & & \highlight{$6.56$\sstd{$0.30$}} & $36.18$\sstd{$1.51$} & \highlight{$82.66$\sstd{$1.06$}} & $70.58$\sstd{$0.20$} \\
\midrule
MLE & \multirow{2}{*}{\rotatebox[origin=c]{90}{Point}} &$7.19$\sstd{$0.27$} & \highlight{$7.88$\sstd{$1.92$}} & $76.19$\sstd{$3.54$} & \highlight{$79.05$\sstd{$3.91$}} \\
MAP & &$7.18$\sstd{$0.36$} & \highlight{$7.71$\sstd{$1.31$}} & $79.82$\sstd{$1.10$} & \highlight{$78.75$\sstd{$3.32$}} \\
\bottomrule
    \end{tabular}
    \vspace{-1mm}
    \caption{\textbf{Tabular Experiments}. We test the generalization ability of different in-context estimators to perform well zero-shot to a suite of real-world tabular tasks under the different assumed probabilistic models.}
        \vspace{-5mm}
    \label{tab:tabular}
\end{table}
\begin{table*}[t]
    \centering
    \small
    % \scriptsize
    % \def\arraystretch{1.5}
    \setlength{\tabcolsep}{3 pt}
    % \resizebox{\linewidth}{!}{
    \begin{tabular}{@{}l r cc cc cc cc}
        \toprule
          & & \multicolumn{4}{c}{\textit{$L_2$ Loss} ($\downarrow$)} & \multicolumn{4}{c}{\textit{Accuracy} ($\uparrow$)}\\
         \cmidrule(lr){3-6}\cmidrule(lr){7-10}
        & \textbf{Objective} & \multicolumn{4}{c}{\textbf{NLR}} & \multicolumn{4}{c}{\textbf{NLC}} \\
        \cmidrule(lr){3-10}
        & & \multicolumn{2}{c}{\textsc{TanH}} & \multicolumn{2}{c}{\textsc{ReLU}} & \multicolumn{2}{c}{\textsc{TanH}} & \multicolumn{2}{c}{\textsc{ReLU}} \\
        \cmidrule(lr){3-10}
        & & $1$D & $50$D & $1$D & $50$D & $2$D $5$cl & $50$D $5$cl & $2$D $5$cl & $50$D $5$cl \\
\midrule
\multirow{3}{*}{Baseline} & Random & $26.02$\sstd{$0.43$} & $29.2$\sstd{$0.1$} & $514.40$\sstd{$3.71$} & $15106.4$\sstd{$110.3$} & $19.8$\sstd{$0.8$} & $20.0$\sstd{$0.3$} & $19.0$\sstd{$1.1$} & $19.5$\sstd{$0.6$} \\
& Optimization & $0.56$\sstd{$0.01$} & $27.4$\sstd{$0.1$} & $2.46$\sstd{$0.09$} & $4723.8$\sstd{$7.9$} & $89.4$\sstd{$0.1$} & $35.2$\sstd{$0.1$} & $94.2$\sstd{$0.1$} & $61.0$\sstd{$0.1$} \\
& Langevin & $0.34$\sstd{$0.01$} & $36.3$\sstd{$0.4$} & \textsc{N/A} & \textsc{N/A} & $84.0$\sstd{$0.3$} & $26.8$\sstd{$0.3$} & $91.9$\sstd{$0.3$} & $50.1$\sstd{$0.3$} \\
& HMC & $0.65$\sstd{$0.01$} & $32.7$\sstd{$0.4$} & $8.16$\sstd{$0.48$} & $12899.8$\sstd{$11.7$} & $75.2$\sstd{$0.3$} & $25.3$\sstd{$0.4$} & $79.6$\sstd{$0.4$} & $51.3$\sstd{$0.5$} \\
& Langevin-multiple & $0.31$\sstd{$0.00$} & $20.3$\sstd{$0.0$} & \textsc{N/A} & \textsc{N/A} & $87.6$\sstd{$0.1$} & $35.0$\sstd{$0.2$} & $94.3$\sstd{$0.1$} & $61.7$\sstd{$0.1$} \\
& HMC-multiple & $0.63$\sstd{$0.00$} & $23.6$\sstd{$0.1$} & $7.98$\sstd{$0.14$} & $12752.5$\sstd{$4.2$} & $77.4$\sstd{$0.1$} & $33.3$\sstd{$0.5$} & $81.7$\sstd{$0.1$} & $60.2$\sstd{$0.1$} \\
\midrule
\multirow{3}{*}{Gaussian} & Fwd-KL &$25.96$\sstd{$0.45$} & $29.2$\sstd{$0.1$} & $276.37$\sstd{$11.57$} & $8422.3$\sstd{$77.1$} & $20.5$\sstd{$0.8$} & $20.5$\sstd{$0.3$} & $50.8$\sstd{$1.4$} & $49.4$\sstd{$0.6$} \\
& Rev-KL &$1.96$\sstd{$0.54$} & $24.1$\sstd{$0.6$} & $8.72$\sstd{$1.51$} & $4756.4$\sstd{$34.9$} & $19.9$\sstd{$0.8$} & $20.0$\sstd{$0.4$} & $61.2$\sstd{$0.8$} & $29.6$\sstd{$0.2$} \\
& Sym-KL &$3.40$\sstd{$0.07$} & \highlight{$20.4$\sstd{$0.0$}} & $11.15$\sstd{$2.27$} & \highlight{$4569.2$\sstd{$26.9$}} & $20.1$\sstd{$0.8$} & $20.2$\sstd{$0.4$} & $62.0$\sstd{$0.6$} & $28.4$\sstd{$0.4$} \\
\midrule
\multirow{3}{*}{Norm. Flows} & Fwd-KL &$25.68$\sstd{$0.58$} & $29.2$\sstd{$0.1$} & $244.92$\sstd{$8.60$} & $7702.5$\sstd{$70.3$} & $20.6$\sstd{$1.2$} & $20.8$\sstd{$0.4$} & $54.4$\sstd{$0.4$} & $50.5$\sstd{$0.7$} \\
& Rev-KL &$2.93$\sstd{$0.02$} & $23.3$\sstd{$0.3$} & \highlight{$6.14$\sstd{$0.42$}} & $4791.8$\sstd{$42.9$} & $20.3$\sstd{$1.5$} & $20.1$\sstd{$0.4$} & $60.8$\sstd{$0.6$} & $55.6$\sstd{$0.5$} \\
& Sym-KL &$2.08$\sstd{$0.42$} & $21.8$\sstd{$0.8$} & $14.96$\sstd{$7.22$} & $4868.0$\sstd{$188.8$} & $20.7$\sstd{$1.2$} & $20.1$\sstd{$0.2$} & $63.6$\sstd{$0.4$} & $56.5$\sstd{$0.1$} \\
\midrule
\multirow{3}{*}{Diffusion} & Score-Based &$27.24$\sstd{$0.84$} & $30.3$\sstd{$0.1$} & $341.06$\sstd{$43.24$} & $10640.5$\sstd{$290.9$} & $20.4$\sstd{$1.0$} & $19.9$\sstd{$0.6$} & $38.6$\sstd{$3.2$} & $38.5$\sstd{$1.1$} \\
& Flow-Matching &$25.63$\sstd{$0.49$} & $29.4$\sstd{$0.1$} & $261.80$\sstd{$9.77$} & $8480.2$\sstd{$177.6$} & $19.5$\sstd{$0.5$} & $20.0$\sstd{$0.3$} & $52.4$\sstd{$3.1$} & $48.9$\sstd{$0.4$} \\
& pDEM &$26.80$\sstd{$0.98$} & $28.3$\sstd{$0.0$} & $729.49$\sstd{$95.99$} & $9646.4$\sstd{$542.5$} & $20.4$\sstd{$0.9$} & $20.0$\sstd{$0.4$} & $59.1$\sstd{$0.7$} & $55.6$\sstd{$0.1$} \\
\midrule
\multirow{2}{*}{Point} & MLE & \highlight{$0.53$\sstd{$0.04$}} & $25.6$\sstd{$0.3$} & \highlight{$5.98$\sstd{$0.50$}} & $4823.3$\sstd{$43.3$} & \highlight{$85.6$\sstd{$1.1$}} & \highlight{$36.6$\sstd{$0.2$}} & \highlight{$91.2$\sstd{$1.6$}} & $58.9$\sstd{$0.3$} \\
& MAP & \highlight{$0.54$\sstd{$0.04$}} & $25.1$\sstd{$0.2$} & \highlight{$6.83$\sstd{$0.96$}} & $4869.7$\sstd{$116.1$} & $20.3$\sstd{$1.4$} & $20.1$\sstd{$0.6$} & $85.4$\sstd{$0.3$} & \highlight{$59.3$\sstd{$0.1$}} \\
\bottomrule
    \end{tabular}
    % }
    \caption{Comparison of various in-context estimators in inferring the parameters of a $2$-layered neural network for nonlinear regression and classification tasks of varying dimensionalities, number of classes and activation functions. Amortized point estimators considerably outperform posterior counterparts, especially for high-dimensional classification tasks.}
    \vspace{-5mm}
    \label{tab:variable_dim_2l_ens}
\end{table*}
\section{Additional Experiments}
\label{apdx:additional_experiments}

\subsection{In-distribution evaluation of in-context estimators}
\label{apdx:in-distribution}

\subsubsection{Fixed-Dimensional}
\label{apdx:fixed-dim}
In this section, we first outline the different setups for each probabilistic model that serve as part of our empirical analysis. We note that for each probabilistic model as well as each configuration of the said model, a different in-context estimator needs to be learned.

\textbf{GM}: For estimating the mean of a Gaussian distribution, we train different in-context estimators for a $2$-dimensional and $100$-dimensional problem, yielding $2$ tasks.

\textbf{GMM}: For modeling the means of a mixture of Gaussian distribution, different in-context estimators are trained for $2$ and $5$ dimensional observations, with $2$ and $5$ underlying number of clusters. This results in $4$ tasks.

\textbf{LR}: To estimate the weights of a linear regression model, we consider $1$ and $100$ dimensional observations with a scalar target in each setting, resulting in $2$ tasks. 
 
\textbf{NLR}: This experiment requires estimating the parameters of a neural network, where the observations can be $1$ or $25$ dimensional, the number of hidden layers $1$ or $2$, and the activation function can be \textsc{TanH} or \textsc{ReLU}. In total this leads to $8$ different tasks.

\textbf{LC}: To estimate the weights of a linear classifier, we consider $2$ and $100$ dimensional observations with $2$ and $5$ classes, resulting in $4$ tasks.

\textbf{NLC}: Similar to NLR, we consider $2$ or $25$ dimensional observations, $1$ or $2$ number of hidden layers, \textsc{TanH} or \textsc{ReLU} activation function and $2$ or $5$ number of classes, leading to $16$ different tasks.

In total, this leads to a total of $36$ tasks, where every in-context learner is separately trained for each of the tasks. Additionally, for each task and amortized estimator, we conduct our investigation using $6$ seeds leading to $36 \times 6$ models being trained for each estimator.

\cref{tab:fixed_dim} highlights the single-sample performance metrics on high-dimensional tasks for each of the class of probabilistic models. The results corresponding to other configurations of the probabilistic models are abstracted away in \cref{fig:rank}.

\subsubsection{Variable-Dimensional}
\label{apdx:variable-dim}
Next, we look at inferring parameters for variable number of features, and outline the different setups for each probabilistic model. Note that for each assumed model, a single estimator is trained to solve various input number of features.

\textbf{GM}: We evaluate on estimating the mean of a Gaussian distribution for $2$, $50$ and $100$ dimensional problem, leading to $3$ tasks with only $1$ model trained for all of them.

\textbf{GMM}: We consider $2$ and $5$ dimensional observations, with $2$ and $5$ underlying number of clusters. This results in $4$ tasks, with $2$ different models being trained corresponding to the different number of clusters.

\textbf{LR}: The observations are $1$, $50$ and $100$ dimensional with a scalar target, resulting in $3$ tasks and only $1$ trained model. 
 
\textbf{NLR}: Here, the observations can be $1$, $50$ or $100$ dimensional, the number of hidden layers $1$ or $2$, and the activation function can be \textsc{TanH} or \textsc{ReLU}. In total this leads to $12$ different tasks, with $4$ different models trained.

\textbf{LC}: We consider $2$, $50$ and $100$ dimensional observations with $2$ and $5$ classes, resulting in $6$ tasks and $2$ models.

\textbf{NLC}: We evaluate $2$, $50$ and $100$ dimensional observations, $1$ or $2$ number of hidden layers, \textsc{TanH} or \textsc{ReLU} activation function and $2$ or $5$ number of classes, leading to $24$ different tasks and $8$ different trained models.

In total, this leads to a total of $52$ tasks, where the in-context estimators generalize across input dimensions. Given $6$ seeds for each analysis, this leads to $18 \times 6$ models being trained for each estimator.

\cref{tab:variable_dim} highlights the single-sample performance metrics on high-dimensional tasks for each probabilistic model. Additionally \cref{tab:variable_dim_2l_ens} highlights the ensemble-based predictive metrics on considerably harder problems, \ie estimating the parameters of a 2-layered neural network. The results corresponding to other configurations of the probabilistic models are abstracted away in \cref{fig:rank}.

\subsection{Misspecification}

\subsubsection{Synthetic}
\label{apdx:misspecification}
We consider three different data-generating families: linear regression (LR), nonlinear regression modeled through a single layered neural network with \textsc{TanH} activation function (NLR), and Gaussian Process (GP) with the radial basis function kernel. Given a pair of data-generating processes $(\chi_{\rm train}, \chi_{\rm test})$, we train in-context estimators on $\chi_{\rm train}$ and then evaluate them on $\chi_{\rm test}$. The underlying modeling assumption is in line with $\chi_{\rm train}$. For example, if $\chi_{\rm train}$ is LR and $\chi_{\rm test}$ is GP, then the likelihood function is set as the LR probabilistic model described in \cref{apdx:probabilistic_models}. This allows us to train sample-based methods as well, and then evaluate them in OoD scenarios.

Since variational methods and point estimators can be trained on data different from the modeling assumption $p$, we also consider a setting where they are trained directly on $\gD \sim \chi_{\rm test}$, assuming that it does not expose $\theta$ and thus one cannot simply change the modeling assumption. This is referred to as ``+ switched data" in the results. We refer the reader to \cref{tab:misspecification} for additional results corresponding to the single-sample metrics.

\subsubsection{Tabular}
\label{apdx:tabular}
For tabular regression tasks, we use the \textit{OpenML-CTR23} benchmarking suite~\citep{fischer2023openmlctr23}, applying a filtering process to exclude datasets with over 2000 examples, more than 100 features, or missing values (NaNs). Similarly, for classification, we utilize the \textit{OpenML-CC18} benchmark~\citep{bischl2019openmlcc18}, applying the same filtering criteria while additionally removing datasets that are not binary classification problems. This process results in a final selection of 9 regression and 13 classification datasets. The selected datasets are:

Regression: \textsc{airfoil\_self\_noise}, \textsc{concrete\_compressive\_strength}, \textsc{energy\_efficiency}, \textsc{solar\_flare}, \textsc{student\_performance\_por}, \textsc{QSAR\_fish\_toxicity}, \textsc{red\_wine}, \textsc{socmob}, and \textsc{cars}.

Classification: \textsc{credit-g}, \textsc{diabetes}, \textsc{tic-tac-toe}, \textsc{pc4}, \textsc{pc3}, \textsc{kc2}, \textsc{pc1}, \textsc{banknote-authentication}, \textsc{blood-transfusion-service-center}, \textsc{ilpd}, \textsc{qsar-biodeg}, \textsc{wdbc}, and \textsc{climate-model-simulation-crashes}.

For each of the datasets, we evaluate using a $5$-fold cross validation and use $6$ seeds. The inputs to the in-context estimators are normalized to have zero-mean and unit-variance. Evaluation of different estimators through the single-sample metrics for tabular tasks is provided in \cref{tab:tabular}.
\end{document}